\documentclass[11pt]{article}

\usepackage[T1]{fontenc}
\usepackage[utf8]{inputenc}
\usepackage{lmodern}
\usepackage{textcomp}
\usepackage[a4paper,margin=2.5cm]{geometry}
\usepackage{amsmath,amssymb}
\usepackage{bm}
\usepackage{graphicx}
\usepackage{longtable,booktabs,array,multirow,calc}
\usepackage{caption}
\usepackage{microtype}
\usepackage{enumitem}
\usepackage{indentfirst}
\setlist{itemsep=2pt plus 1pt, parsep=0pt, topsep=4pt plus 2pt, partopsep=0pt}
\setlist[enumerate,1]{label=\textbf{(\arabic*)}, labelindent=\parindent, leftmargin=*, align=left, widest=10}
\setlist[enumerate,2]{label=(\alph*), labelindent=0pt, leftmargin=*, align=left}
\newlist{steps}{enumerate}{2}
\setlist[steps,1]{label=\arabic*), labelindent=2\parindent, leftmargin=*, align=left, widest=10}
\setlist[steps,2]{label=(\alph*), labelindent=0pt, leftmargin=*, align=left}
\usepackage[numbers,sort&compress]{natbib}
\usepackage{xurl}
\usepackage[hidelinks]{hyperref}

\usepackage{etoolbox}
\AtBeginEnvironment{longtable}{\footnotesize\setlength{\tabcolsep}{3pt}}

\title{A Time-Frequency-Condition Multimodal Fusion Foundation Model for Universal Rotating-Machinery Fault Diagnosis under Multi-Element Heterogeneous Scenarios}

\author{Xinyu Zou\textsuperscript{a}, Jian Ma\textsuperscript{a,b}, Liang Ma\textsuperscript{c}, Chen Lu\textsuperscript{a,b}, Tao Peng\textsuperscript{d}, Yingjie Li\textsuperscript{a,*}\\[1ex]
{\small\itshape \textsuperscript{a} Hangzhou International Innovation Institute, Beihang University, China}\\
{\small\itshape \textsuperscript{b} School of Reliability and Systems Engineering, Beihang University, China}\\
{\small\itshape \textsuperscript{c} Beijing Institute of Spacecraft System Engineering, China}\\
{\small\itshape \textsuperscript{d} School of Automation, Central South University, China}\\
{\small\itshape \textsuperscript{*} Corresponding author. E-mail: yingjieli@buaa.edu.cn}}

\date{}

\begin{document}

\maketitle

\begin{abstract}
Intelligent fault diagnosis models are typically developed for a single
machine under a specific set of operating conditions, such that each new
machine or condition requires retraining from scratch. This
one-machine-one-model paradigm scales poorly. Although transfer learning
and domain generalization can reduce the labeling burden, they primarily
seek domain-invariant commonality, potentially discarding
machine-specific characteristics that are critical for diagnosis. This
paper proposes TFC-MMFD, a time-frequency-condition multimodal fusion
foundation model for rotating-machinery fault diagnosis under
multi-element heterogeneous scenarios. Statistical indicators,
time-frequency images, vibration sequences, and textual machine and
condition descriptions are tokenized into a unified language-like space,
allowing inputs of different physical origin to be represented as a
single sequence. A LLaMA backbone, together with cross-modal
bidirectional attention and a Q-Former, then distills this sequence into
fault-mode queries. Diagnosis is performed by matching these queries
against natural-language fault-mode anchors rather than fixed class
indices, while regression-adversarial condition disentanglement and
condition-conditioned masked pre-training improve robustness to changing
conditions. Across four progressive tasks on ten public bearing datasets
spanning diverse bearing types, speeds, and loads, TFC-MMFD attains
99.40\% accuracy and 99.16\% macro-F1, maintains over 80\% accuracy on
most interpolated unseen conditions without adaptation, and improves
from 20.26\% to 95.63\% accuracy on an unseen machine with only five
labeled samples per class. It outperforms the strongest of eight
domain-generalization baselines by 10.57 percentage points and compares
favorably with a recent bearing foundation model while using less than
half of its fine-tuning data, indicating that a single pre-trained model
can be efficiently redeployed across machines at low cost.
\end{abstract}


\section{Introduction}\label{sec:1}

Electromechanical systems are core components of industrial equipment,
and their reliability directly affects functional performance,
operational safety, and mission success. As these systems become more
complex, faults are more difficult and costly to manage. Prognostics and
Health Management (PHM) supports condition-based and preventive
maintenance by combining sensor data with diagnostic and prognostic
algorithms. It enables timely assessment of equipment health, reduces
downtime, and improves availability~\cite{ref1,ref2,ref3}. Increasing automation,
connectivity, and deployment scale have created a growing need for
diagnostic methods that are scalable, transferable, and suitable for
distributed applications.

Machine-learning and deep-learning methods have improved the efficiency
and accuracy of fault detection and diagnosis~\cite{ref4,ref5,ref6}. In industrial
applications, however, many methods remain task-dependent and generalize
poorly. A model developed for one machine or data condition may perform
differently on another, which often requires a separate design and
training process for each application. Performance can also decline when
operating conditions, environments, or data quality shift beyond the
training distribution. Together, the diversity of equipment, diagnostic
tasks, and design factors increases the time and cost required to
develop effective PHM systems.

\subsection{Related work on cross-domain diagnosis and foundation models}\label{sec:1.1}

\textbf{Transfer learning and domain generalization.} To improve
adaptability under data-limited, machine-diverse, and
environment-varying scenarios, transfer learning and domain
generalization play important roles in industrial fault diagnosis.
Transfer learning transfers knowledge from one or more source domains
(specific machines or conditions) to a target domain (different machines
or unseen conditions), even when target data are scarce, reducing the
cost of building new diagnosis models~\cite{ref7,ref8,ref9}. Domain generalization
advances this idea further by training a model to recognize common fault
patterns and features across multiple domains, thereby enabling accurate
diagnosis for previously unseen environments or machines without any
target-domain data~\cite{ref10}. Domain-adaptation methods are commonly
grouped into instance transfer, model transfer, and feature transfer.
Feature transfer is the most common and is realized either through
discrepancy metrics such as Maximum Mean Discrepancy (MMD)~\cite{ref11} or
through adversarial learning, represented by the Domain-Adversarial
Neural Network (DANN)~\cite{ref12}. Since 2020, domain-generalization
research has expanded rapidly, branching into open-set, imbalanced, and
federated variants~\cite{ref13,ref14,ref15}, and moving toward more realistic
deployment settings~\cite{ref16,ref17,ref18}.

Despite their potential, both transfer learning and domain
generalization have limitations. Transfer learning depends on the
similarity between the source and target domains; a large domain gap can
cause negative transfer, and selecting an appropriate source domain and
transfer strategy often requires manual analysis~\cite{ref19}. Domain
generalization methods seek common fault representations across domains,
but some domain differences cannot be removed without losing useful
diagnostic information. Features associated with a particular speed or
load may still be necessary to distinguish similar faults. Recent work
has therefore moved beyond strict domain invariance and has begun to
retain object-, condition-, and background-specific information to
assist cross-domain diagnosis~\cite{ref20,ref21,ref22,ref23}.

\textbf{Universal foundation models and relevant techniques.} Foundation
models in computer vision and natural language processing provide
inspiring insights for general-purpose fault-diagnosis modeling~\cite{ref24}. Large language models~\cite{ref25}, including the LLaMA series~\cite{ref26,ref27}, and vision models such as SAM~\cite{ref28} and CLIP~\cite{ref29}
support cross-task and cross-domain representation learning. Three
techniques are particularly relevant: Transformer-based feature
extraction~\cite{ref30} for capturing long-range dependencies in monitoring
signals; self-supervised learning, including masked modeling,
autoregression, autoencoding, and contrastive learning~\cite{ref31,ref32,ref33}, for
learning from unlabeled data; and intermediate multimodal fusion with
Transformer cross-attention~\cite{ref34,ref35,ref36}. These techniques have also
proven effective in fault-diagnosis applications~\cite{ref37,ref38,ref39}.

\subsection{Challenges of universal fault-diagnosis modeling under multi-element heterogeneous scenarios}\label{sec:1.2}

Compared with vision and language, cross-domain universal modeling in
fault diagnosis presents more complex characteristics and requirements:
the diversity of machines and environments, the evolution of system
state and parameter distributions, and the balance between customization
and universality. A cross-domain PHM foundation model must therefore
accommodate multi-dimensional, multi-element differentiated scenarios,
i.e., multi-element heterogeneous scenarios. These scenarios comprise
object heterogeneity (differences in the kind and nature of machines,
e.g., different models of the same object type or entirely different
object types), data heterogeneity (diversity in the source, type, and
quality of data and features), and task heterogeneity (diversity of the
task types and requirements the diagnosis model must handle).

Adapting foundation-model techniques from language and vision to fault
diagnosis raises challenges in both data fusion and model training.
First, fault-diagnosis data include raw sensor signals, derived
features, operating conditions, and machine information. These inputs
differ in format, scale, and physical meaning, so they require a
consistent representation and alignment strategy. Second, industrial
monitoring signals are low-level and often contain sparse fault
information. Self-supervised tasks must therefore be designed for these
signals, while the training process must separate background and
operating-condition effects from fault-related information. This paper
focuses on object and data heterogeneity and addresses them through
unified representation, architecture design, and training strategy.

\subsection{Motivation and contributions}\label{sec:1.3}

Universal fault diagnosis in heterogeneous settings requires coordinated
treatment of data representation, model architecture, and training.
Time-domain signals, frequency- and time-frequency-domain features, and
object and operating-condition information differ in form, dimension,
and physical meaning. They must therefore be represented consistently
without removing cues that remain useful for diagnosis. The architecture
must also convert sparse industrial signals with low information density
into dense representations suitable for a language-model backbone.
Finally, the training objectives must disentangle the influence of
continuously varying operating conditions while allowing the diagnostic
output space to remain semantically interpretable and extensible.
Motivated by these observations, this paper proposes the
Time-Frequency-Condition Multimodal Fault Diagnosis model (TFC-MMFD),
which addresses the challenges within a single end-to-end framework. The
main contributions are as follows:

First, a TFC multimodal fusion framework is developed to represent
heterogeneous inputs in a common language-like space. Time-domain
signals, frequency- and time-frequency-domain features, and object and
operating-condition information are organized as numerical scalars,
text, images, and sequences. Scalars are expanded with a
polynomial-Chebyshev-trigonometric operator, while the other inputs are
encoded with pre-trained CLIP text, CLIP vision, and Wav2Vec2 encoders.
Treating condition and structural information as an explicit modality
allows a single model to learn jointly from heterogeneous machines and
multi-source datasets.

Second, a hierarchical multimodal architecture is designed to process
sparse, low-information-density industrial signals. By integrating a
multimodal space mapping module, a cross-modal bidirectional attention
module, a LLaMA backbone fine-tuned with Low-Rank Adaptation (LoRA), and
a Q-Former query module, the architecture progressively maps, aligns,
and fuses the heterogeneous inputs into high-level semantic fault
representations. This design bridges the semantic gap between raw
vibration data and the language space while maintaining computational
feasibility.

Third, a multi-task training strategy is developed to ensure
generalization and extensibility. Specifically, regression-adversarial
training replaces the discrete domain classifier with a continuous
regressor via a gradient reversal layer, extending condition
disentanglement to unseen conditions. In addition, a masked pre-training
method is used to reconcile the conflict between condition invariance
and accurate reconstruction. Furthermore, a contrastive learning method
formulates diagnosis as matching between fault features and text-based
fault-mode anchors. The anchors provide semantic labels within a shared
output space and can be extended by adding fault descriptions.

\begin{figure}[htbp]
\centering
\includegraphics[width=0.87\textwidth]{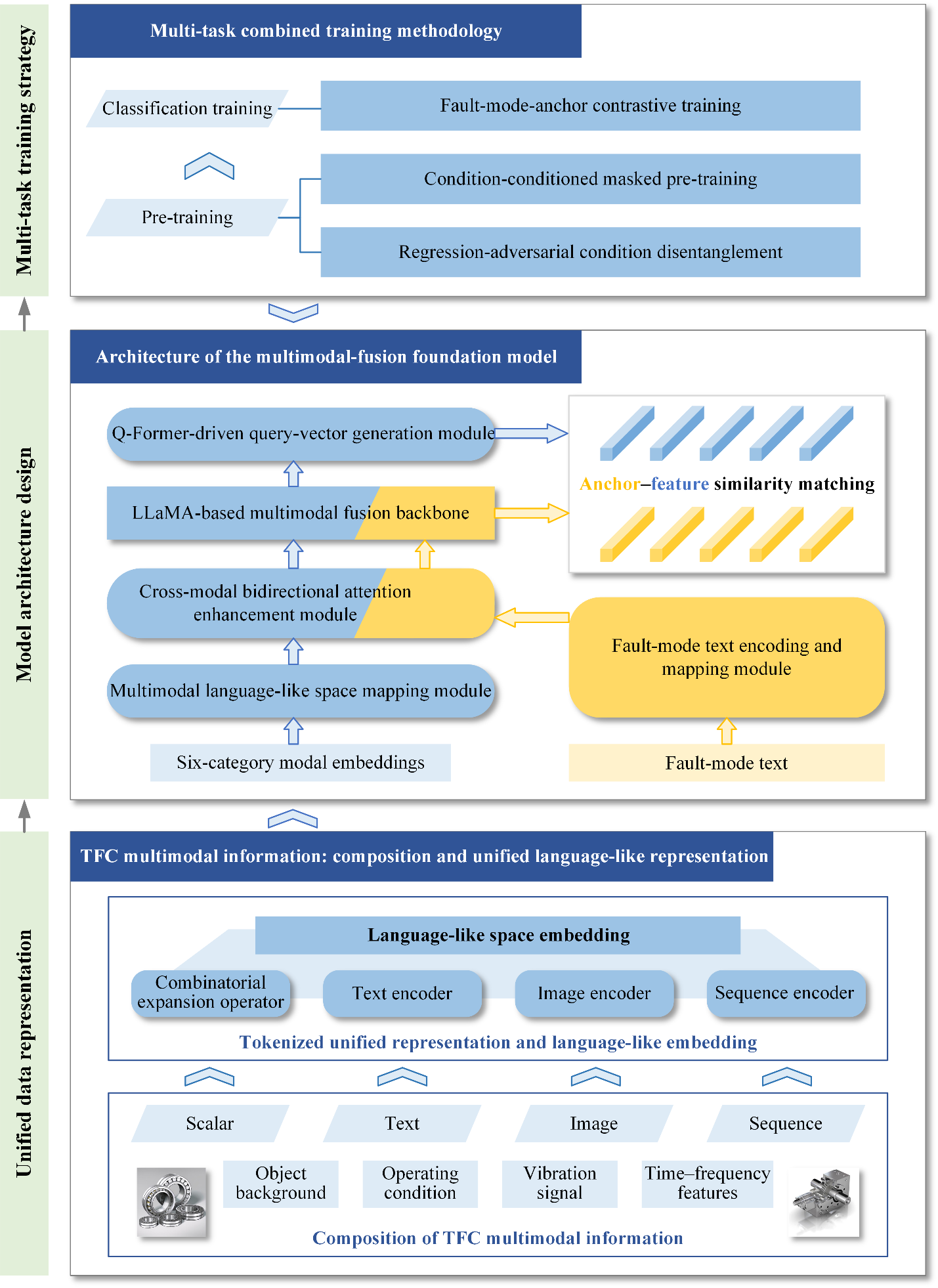}
\caption{Overall technical route of the TFC multimodal fusion fault-diagnosis foundation model.}\label{fig:1}
\end{figure}

The proposed method is evaluated on ten public bearing datasets across
four tasks: universal diagnosis, cross-condition zero-shot
generalization, cross-machine zero-shot generalization, and
cross-machine few-shot fine-tuning. TFC-MMFD achieves 99.40\% accuracy
and a 99.16\% macro-F1 on the mixed benchmark and demonstrates reliable
zero-shot generalization to interpolated unseen conditions. On the PU
dataset, the cross-condition accuracy surpasses the best of eight
domain-generalization baselines by 10.57 percentage points. For
cross-machine adaptation, three- and five-shot fine-tuning substantially
improves performance. For example, accuracy on the XJTU dataset
increases from 20.26\% to 95.63\% in the five-shot setting. Notably,
TFC-MMFD compares favorably with a recent billion-scale bearing
foundation model while using less than half of its fine-tuning data.

The remainder of this paper is organized as follows. Section~\ref{sec:2} defines
the composition of the TFC multimodal information and presents its
tokenized unified language-like representation. Section~\ref{sec:3} details the
architecture of the multimodal-fusion fault-diagnosis foundation model.
Section~\ref{sec:4} describes the multi-task combined training methodology.
Section~\ref{sec:5} reports the case study and comparative analyses, and Section~\ref{sec:6}
concludes the paper with limitations and future work. The overall
technical route is shown in Fig.~\ref{fig:1}.

\section{Time-Frequency-Condition Multimodal Information and Its Tokenized Unified Language-like Representation}\label{sec:2}

In modern industrial equipment, faults naturally span multiple physical
domains, manifesting as transient impacts in the time domain, harmonics
or sidebands in the frequency domain, and multivariable responses
coupled with machine structure and operating conditions. A single
feature representation may not capture this diverse diagnostic
information, leading to low accuracy and poor generalization under
various operating conditions. To address these challenges, TFC
multimodal fusion jointly integrates time-domain signals, time-frequency
features, and condition and structural information, thereby improving
the generalization and robustness of the foundation model.

Conventional concatenation or sequential fusion struggles to capture
fine-grained associations across modalities. Inspired by natural
language processing, we introduce tokenization and vector-space
embedding to bridge this gap. By converting features from different
physical domains into token-like representations mapped into a unified
language-like space, the model can capture complex inter-modal
dependencies and process multimodal signals efficiently.

Toward this goal, this section introduces a tokenized unified
representation and embedding mechanism for the TFC multimodal
information. By unifying heterogeneous signals into a serialized
representation, this scheme enables the modalities to be processed
jointly while retaining their distinct diagnostic information.

\subsection{Composition of the time-frequency-condition (TFC) multimodal information}\label{sec:2.1}

\subsubsection{Classical time-, frequency-, and time-frequency-domain feature extraction}\label{sec:2.1.1}

For a single-channel one-dimensional vibration sequence \(x_{i}\) of
length \(L_{x}\), indicators and image-form features that characterize
faults are extracted at three levels: the time, frequency, and
time-frequency domains. In the time domain, ten common statistics are
adopted: mean, root-mean-square (RMS), variance, skewness, kurtosis,
peak, peak-to-peak, crest factor, impulse factor, and margin factor
(Table~\ref{tab:1}). These statistics quantify the overall level, symmetry, and
sharpness of the amplitude distribution and provide an important basis
for identifying impulsive or transient-impact faults. Four extended
indicators (Shannon entropy, approximate entropy, the Hurst exponent,
and the Lyapunov exponent) are introduced to further characterize the
complexity and nonlinear evolution of the sequence, since such
non-Gaussian, non-stationary characteristics are often closely
associated with specific mechanical fault modes.

In frequency-domain analysis, eight common features are used. The
dominant frequency, center frequency, and bandwidth indicate where the
main energy and harmonics concentrate, helping to distinguish the
characteristic bands of different fault types. Envelope spectra and
their statistics are further introduced because of the physical
mechanism of rolling-bearing faults: a localized defect generates
periodic impacts that modulate the high-frequency structural resonances,
and envelope analysis demodulates these impacts so that fault-related
components emerge clearly at the fault characteristic frequencies.
Envelope-spectrum features are therefore particularly informative for
exposing frequency anomalies in bearing diagnosis~\cite{ref40}.

In time-frequency analysis, classical short-time Fourier transform
(STFT) and wavelet transform generate image-form representations of
time-varying features, providing interpretable fault signatures across
multiple scales and resolutions. From these images, mean amplitude,
maximum amplitude, and wavelet energy are further extracted to help
locate the timing and frequency band of transient impacts and the
temporal evolution of fault energy. In total, three time-frequency image
features and five time-frequency statistical features are extracted
(Table~\ref{tab:1}).

\subsubsection{Object-background and operating-condition information as the ``Condition'' modality}\label{sec:2.1.2}

This study focuses on rotating-machinery diagnosis. The term
``Condition'' (the C in TFC) includes both operating-condition
information and object-background information, such as machine
structure, type, and physical parameters. Rolling bearings, which are
widely monitored components in rotating machinery, serve as the
representative object class in the experimental validation. For each
bearing, the model uses the bearing type (e.g., ball or roller bearing),
pitch diameter, number of rolling elements, rolling-element diameter,
and contact angle. These geometric parameters are also used to calculate
the fault characteristic frequencies: ball pass frequency on the inner
race (BPFI), ball pass frequency on the outer race (BPFO), ball spin
frequency (BSF), and fundamental train frequency (FTF). These
frequencies serve as physics-based prior knowledge that links machine
structure to the expected spectral signatures of faults. This
information helps the model represent the same fault mode consistently
across machines with different structures, which is important for
universal diagnosis.

Operating-condition information includes the rotational frequency, load
level, and sampling frequency under different operating states. These
quantities vary across time and tasks and exert a substantial influence
on the amplitude, frequency content, and noise level of fault signals:
under high-speed, high-load conditions, vibration signals tend to be
more intense and susceptible to external-excitation coupling, whereas
under low-load or quasi-steady conditions some fault features may be
masked or weakened. Incorporating machine-object information and
operating conditions into a single framework allows the feature analysis
to be adapted to specific condition changes, which is crucial for a
universal foundation model applicable to a variety of rotating
machinery.

Table~\ref{tab:1} summarizes the composition of all TFC multimodal information
used in this paper, and Table~\ref{tab:2} gives the specific object-background
information for bearings.

\begin{longtable}[]{@{}
  >{\raggedright\arraybackslash}p{(\columnwidth - 8\tabcolsep) * \real{0.1450}}
  >{\raggedright\arraybackslash}p{(\columnwidth - 8\tabcolsep) * \real{0.2000}}
  >{\raggedright\arraybackslash}p{(\columnwidth - 8\tabcolsep) * \real{0.1900}}
  >{\raggedright\arraybackslash}p{(\columnwidth - 8\tabcolsep) * \real{0.1050}}
  >{\raggedright\arraybackslash}p{(\columnwidth - 8\tabcolsep) * \real{0.3600}}@{}}
\caption{Composition of the TFC multimodal information.}\label{tab:1}\\
\toprule\noalign{}
\begin{minipage}[b]{\linewidth}\raggedright
\textbf{Category}
\end{minipage} & \begin{minipage}[b]{\linewidth}\raggedright
\textbf{Name}
\end{minipage} & \begin{minipage}[b]{\linewidth}\raggedright
\textbf{Symbol}
\end{minipage} & \begin{minipage}[b]{\linewidth}\raggedright
\textbf{Data type}
\end{minipage} & \begin{minipage}[b]{\linewidth}\raggedright
\textbf{Description}
\end{minipage} \\
\midrule\noalign{}
\endhead
\bottomrule\noalign{}
\endlastfoot
Raw signal & Vibration signal & \(x_{v,i}\) & \(\mathbb{R}^{L_{x}}\) &
Single-channel 1-D vibration sequence. \\
\midrule
\multirow{10}{=}{Time-domain features} & Mean & \({TD}_{mean}\) &
\(\mathbb{R}\) & Overall offset level of the signal. \\*
& RMS & \({TD}_{rms}\) & \(\mathbb{R}\) & Energy/effective value,
representing overall amplitude. \\*
& Variance & \({TD}_{var}\) & \(\mathbb{R}\) & Dispersion of the
amplitude distribution. \\*
& Skewness & \({TD}_{skew}\) & \(\mathbb{R}\) & Symmetry of the
distribution; nonzero means asymmetric. \\*
& Kurtosis & \({TD}_{kurt}\) & \(\mathbb{R}\) & Sharpness; especially
sensitive to impulsive faults. \\*
& Peak & \({TD}_{peak}\) & \(\mathbb{R}\) & Maximum value over the
amplitude range. \\*
& Peak-to-peak & \({TD}_{p2p}\) & \(\mathbb{R}\) & Difference between
max and min amplitude. \\*
& Crest factor & \({TD}_{crf}\) & \(\mathbb{R}\) & Ratio of peak to RMS;
relative strength of spikes. \\*
& Impulse factor & \({TD}_{imp}\) & \(\mathbb{R}\) & Ratio of peak to
mean absolute value; impulse-sensitive. \\*
& Margin factor & \({TD}_{mrg}\) & \(\mathbb{R}\) & Ratio of peak to
squared mean of square roots. \\
\midrule
\multirow{4}{=}{Extended time-domain features} & Shannon entropy &
\({TD}_{shen}^{aux}\) & \(\mathbb{R}\) & Uncertainty/randomness of the
sequence. \\*
& Approximate entropy & \({TD}_{apen}^{aux}\) & \(\mathbb{R}\) &
Complexity and rate of new-pattern occurrence. \\*
& Hurst exponent & \({TD}_{hurst}^{aux}\) & \(\mathbb{R}\) &
Persistence/anti-persistence; long-range correlation. \\*
& Lyapunov exponent & \({TD}_{lyap}^{aux}\) & \(\mathbb{R}\) & Whether
the system dynamics are chaotic. \\
\midrule
\multirow{8}{=}{Frequency-domain features} & Dominant frequency &
\({FD}_{main}\) & \(\mathbb{R}\) & Frequency where signal energy
concentrates. \\*
& Center frequency & \({FD}_{center}\) & \(\mathbb{R}\) & Centroid of
the spectral distribution. \\*
& Bandwidth & \({FD}_{bw}\) & \(\mathbb{R}\) & Width of the main
energy/harmonic band. \\*
& Band energy & \({FD}_{energy}\) & \(\mathbb{R}\) & Integrated energy
within a specified band. \\*
& Envelope spectrum & \(P_{envSpec}\) &
\(\mathbb{R}^{L_{H} \times L_{F}}\) & Image from envelope extraction
then spectral analysis. \\*
& Envelope-spectrum mean & \({FD}_{envSpec\_mean}\) & \(\mathbb{R}\) &
Average impact energy across the band. \\*
& Envelope-spectrum max & \({FD}_{envSpec\_max}\) & \(\mathbb{R}\) &
Main-energy/harmonic strength for fault frequencies. \\*
& Envelope-spectrum dominant freq. & \({FD}_{envSpec\_mainF}\) &
\(\mathbb{R}\) & Dominant frequency of the envelope spectrum. \\
\midrule
\multirow{3}{=}{Time-frequency features} & TF amplitude spectrum &
\(P_{ampSpec}\) & \(\mathbb{R}^{L_{T1} \times L_{F1}}\) & STFT amplitude
vs. time-frequency image. \\*
& TF power spectrum & \(P_{pwrSpec}\) &
\(\mathbb{R}^{L_{T2} \times L_{F2}}\) & Power/energy vs. time-frequency
image. \\*
& Wavelet TF map & \(P_{wavTr}\) & \(\mathbb{R}^{L_{T3} \times L_{F3}}\)
& Time-frequency distribution from wavelet transform. \\
\midrule
\multirow{5}{=}{TF statistical features} & Wavelet energy &
\({TF}_{wavE}^{aux}\) & \(\mathbb{R}\) & Energy of wavelet coefficients
across scales. \\*
& Wavelet mean & \({TF}_{wavTr\_mean}^{aux}\) & \(\mathbb{R}\) &
Average of wavelet-transform coefficients. \\*
& Wavelet max & \({TF}_{wavTr\_max}^{aux}\) & \(\mathbb{R}\) & Locates
the most prominent transient impact. \\*
& STFT amplitude mean & \({TF}_{stftMag\_mean}^{aux}\) & \(\mathbb{R}\)
& Average level of the energy distribution. \\*
& STFT amplitude max & \({TF}_{stftMag\_max}^{aux}\) & \(\mathbb{R}\) &
Locates the strongest impact/high-energy component. \\
\midrule
\multirow{4}{=}{Object background} & Object category & \(B_{class}\) &
Text & Category of the monitored object, e.g., rolling bearing
(text). \\*
& Object type & \(B_{type}\) & Text & Subtype, e.g., deep-groove ball or
cylindrical roller bearing (text). \\*
& Object structure & \(B_{stru}\) & \(\mathbb{R}\) & Physical structure
and sizes (e.g., pitch diameter). \\*
& Characteristic frequency & \(B_{freq}\) & \(\mathbb{R}\) & Fault
characteristic / key related frequencies. \\
\midrule
\multirow{3}{=}{Operating condition} & Rotational frequency &
\(W_{rotFreq}\) & \(\mathbb{R}\) & Speed; affects main and sideband
distribution. \\*
& Load level & \(W_{load}\) & \(\mathbb{R}\) & External load;
amplifies/couples fault frequencies. \\*
& Sampling frequency & \(W_{fs}\) & \(\mathbb{R}\) & Sampling frequency
of the acquired signal. \\
\end{longtable}

\begin{longtable}[]{@{}
  >{\raggedright\arraybackslash}p{(\columnwidth - 8\tabcolsep) * \real{0.1140}}
  >{\raggedright\arraybackslash}p{(\columnwidth - 8\tabcolsep) * \real{0.2442}}
  >{\raggedright\arraybackslash}p{(\columnwidth - 8\tabcolsep) * \real{0.1917}}
  >{\raggedright\arraybackslash}p{(\columnwidth - 8\tabcolsep) * \real{0.1125}}
  >{\raggedright\arraybackslash}p{(\columnwidth - 8\tabcolsep) * \real{0.3376}}@{}}
\caption{Object-background information for bearings.}\label{tab:2}\\
\toprule\noalign{}
\begin{minipage}[b]{\linewidth}\raggedright
\textbf{Object}
\end{minipage} & \begin{minipage}[b]{\linewidth}\raggedright
\textbf{Name}
\end{minipage} & \begin{minipage}[b]{\linewidth}\raggedright
\textbf{Symbol}
\end{minipage} & \begin{minipage}[b]{\linewidth}\raggedright
\textbf{Type}
\end{minipage} & \begin{minipage}[b]{\linewidth}\raggedright
\textbf{Description}
\end{minipage} \\
\midrule\noalign{}
\endhead
\bottomrule\noalign{}
\endlastfoot
\multirow{6}{=}{Bearing} & Bearing type & \(B_{bearingType}\) & Text &
Type, e.g., ball or roller bearing. \\*
& Pitch diameter & \(B_{pitchDia}\) & \(\mathbb{R}\) & Pitch-circle
diameter (mm). \\*
& No. of rolling elements & \(B_{rollerNum}\) & \(\mathbb{Z}\) & Number
of rolling elements. \\*
& Rolling-element diameter & \(B_{rollerDia}\) & \(\mathbb{R}\) &
Diameter (mm), for fault-frequency calculation. \\*
& Contact angle & \(B_{contactAngle}\) & \(\mathbb{R}\) & Contact angle
of rolling elements with races. \\*
& Fault characteristic freq. &
\(B_{BPFI}\), \(B_{BPFO}\), \(B_{BSF}\), \(B_{FTF}\) & \(\mathbb{R}\) & BPFI,
BPFO, BSF, FTF. \\
\end{longtable}

\subsection{Tokenized unified representation and language-like space embedding of heterogeneous modalities}\label{sec:2.2}

\begin{figure}[htbp]
\centering
\includegraphics[width=0.69\textwidth]{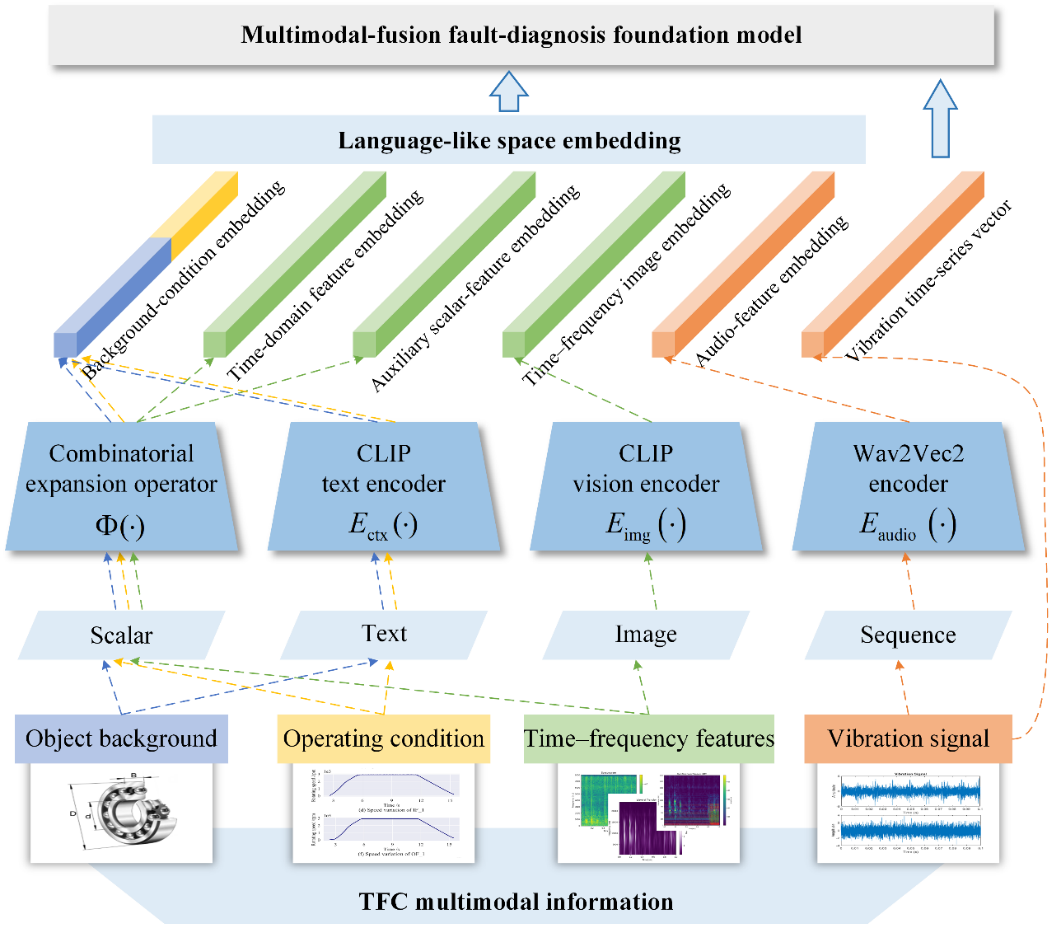}
\caption{Tokenized unified representation and language-like-space embedding of multimodal information.}\label{fig:2}
\end{figure}

The TFC information consists of four data types: numerical scalars,
text, images, and sequences. Each type is mapped to the unified
language-like space using a domain-tailored encoder. The unified
representation allows the modalities to be processed effectively and
provides the token sequence required by the fusion architecture in
Section~\ref{sec:3}. The overall pipeline is illustrated in Fig.~\ref{fig:2}.

\subsubsection{Numerical-scalar embedding via the polynomial-Chebyshev-trigonometric combinatorial expansion operator}\label{sec:2.2.1}

\textbf{(1) Field-level Z-score normalization.} Given the marked
differences in units and magnitude across features, each numerical field
is normalized separately at the field level. For a scalar field
\(x_{j}\mathbb{ \in R}\) with mean \(\mu_{j}\) and standard
deviation \(\sigma_{j}\), the normalization is:

\begin{equation}
{\hat{x}}_{j} = \frac{x_{j} - \mu_{j}}{\sigma_{j}}
\tag{1}\label{eq:1}
\end{equation}

After normalization, all numerical fields are centered at zero with unit
standard deviation, which improves convergence and training stability.

\textbf{(2) Scalar token acquisition.} Unlike images or text, scalars
are usually not processed by a deep encoder. Because of their low
dimensionality, feeding raw scalars directly may cause them to be
underweighted relative to high-dimensional modalities. A ``polynomial
combinatorial expansion'' operator \(\Phi(\cdot)\) is proposed to enrich the scalar
representation. Given a normalized scalar
\(\hat{x}\in\mathbb{R}\), polynomial, Chebyshev, and trigonometric
mappings expand it into a higher-dimensional vector.

\begin{steps}
\item Scaling and hyperbolic-tangent mapping. To prevent gradient explosion
of the polynomial and Chebyshev terms at large values, \(\hat{x}\) is
first scaled by \(\lambda_{scale}\) and passed through tanh(\(\cdot\)):
\begin{equation}
x_{mapped} = \tanh(\lambda_{scale} \cdot \hat{x})
\tag{2}\label{eq:2}
\end{equation}
\item Polynomial features (2nd order):
\begin{equation}
p_{poly} = \lbrack x_{mapped}, {x_{mapped}}^{2}\rbrack
\tag{3}\label{eq:3}
\end{equation}
\item Chebyshev features (3rd order):
\begin{equation}
T_{0} = 1,\quad T_{1} = x_{mapped},\quad T_{2} = 2({x_{mapped}}^{2}) - 1;\quad p_{cheby} = \lbrack T_{0}, T_{1}, T_{2}\rbrack
\tag{4}\label{eq:4}
\end{equation}
\item Sine-cosine features. Let \(freqs = \{ 1, 2, \ldots, K\}\);
the original \(\hat{x}\) (not the tanh-scaled value, since the sine and
cosine functions are bounded) is multiplied by each frequency:
\begin{equation}
{\tilde{x}}_{k} = k \cdot \hat{x},\quad k \in \{ 1, \ldots, K\};\quad p_{\sin} = \lbrack \sin({\tilde{x}}_{1}), \ldots, \sin({\tilde{x}}_{K}), \cos({\tilde{x}}_{1}), \ldots, \cos({\tilde{x}}_{K})\rbrack
\tag{5}\label{eq:5}
\end{equation}
\item Polynomial-feature concatenation. All expanded vectors are
concatenated in turn:
\begin{equation}
\Phi(\hat{x}) = \lbrack \hat{x}, p_{poly}, p_{cheby}, p_{\sin}\rbrack \in \mathbb{R}^{d_{scalar}}
\tag{6}\label{eq:6}
\end{equation}
\end{steps}

\noindent yielding a combinatorial representation in the
polynomial-Chebyshev-trigonometric space; \(\Phi(\hat{x})\) can be fed directly into
the subsequent multimodal Transformer, providing a richer scalar
representation.

\subsubsection{Text embedding via the CLIP text encoder}\label{sec:2.2.2}

For textual data, the CLIP text encoder \(E_{ctx}( \cdot )\) is used~\cite{ref29}. Trained on large-scale image-text contrastive data, it
processes domain terms at the subword level and maps outputs into the
same \(\mathbb{R}^{d}\) space as the image encoder. Let Tokenizer(\(\cdot\))
denote the subword tokenizer and \({Transformer}_{text}( \cdot )\)
denote the internal language model, the encoding s = \(E_{ctx}(S)\)
proceeds as:

\textbf{(1) Tokenization:}

\begin{equation}
\lbrack w_{1}, \ldots, w_{N}\rbrack = \mathrm{Tokenizer}(S),\quad w_{i}\mathcal{ \in V}
\tag{7}\label{eq:7}
\end{equation}

\noindent where \(\mathcal{V}\) is the subword vocabulary and N the subword-sequence length.

\textbf{(2) Word-vector mapping:}

\begin{equation}
e_{i} = \mathrm{EmbeddingMatrix}(w_{i}) \in \mathbb{R}^{d_{text}},\quad i = 1, \ldots, N
\tag{8}\label{eq:8}
\end{equation}

\textbf{(3) Multi-layer attention:}

\begin{equation}
\lbrack h_{1}, \ldots, h_{N}\rbrack = {Transformer}_{text}(e_{1}, \ldots, e_{N})
\tag{9}\label{eq:9}
\end{equation}

\textbf{(4) Global text representation:}

\begin{equation}
s = h_{\lbrack EOT\rbrack} \in \mathbb{R}^{d_{text}}
\tag{10}\label{eq:10}
\end{equation}

The representation at the end-of-text (EOT) token is used as the global
text embedding.

\subsubsection{Time-frequency image embedding via the CLIP vision encoder}\label{sec:2.2.3}

For image information, the CLIP vision encoder \(E_{img}( \cdot )\)
paired with the text encoder is used~\cite{ref29}. The image is divided into
equal-size patches \(p_{j}\); the encoding o = \(E_{img}(P)\) proceeds
as:

\textbf{(1) Linear mapping:}

\begin{equation}
e_{j} = \mathrm{Linear}(p_{j}) \in \mathbb{R}^{d_{img}}
\tag{11}\label{eq:11}
\end{equation}

\textbf{(2) ViT self-attention:}

\begin{equation}
\lbrack h_{\lbrack CLS\rbrack}, h_{1}, \ldots, h_{N_{p}}\rbrack = {Transformer}_{image}(e_{\lbrack CLS\rbrack}, e_{1}, \ldots, e_{N_{p}})
\tag{12}\label{eq:12}
\end{equation}

\textbf{(3) Global image representation:}

\begin{equation}
o = h_{\lbrack CLS\rbrack} \in \mathbb{R}^{d_{img}}
\tag{13}\label{eq:13}
\end{equation}

\subsubsection{Vibration-sequence embedding via the Wav2Vec2 encoder}\label{sec:2.2.4}

Given the strong similarity between vibration waveforms and audio
signals, the Wav2Vec2 encoder \(E_{audio}( \cdot )\) is adopted~\cite{ref41}. Raw signals are resampled to a uniform 16 kHz; for the
resampled single-channel waveform \(x_{v} \in \mathbb{R}^{L_{w}}\),
the encoding a = \(E_{audio}(x_{v})\) proceeds as:

\textbf{(1) Convolutional down-sampling:}

\begin{equation}
\lbrack f_{1}, \ldots, f_{M_{a}}\rbrack = \mathrm{ConvEncoder}(x_{v}),\quad f_{m} \in \mathbb{R}^{d_{f}}
\tag{14}\label{eq:14}
\end{equation}

\textbf{(2) Transformer self-attention:}

\begin{equation}
\lbrack u_{1}, \ldots, u_{M_{a}}\rbrack = {Transformer}_{audio}(f_{1}, \ldots, f_{M_{a}}),\quad u_{m} \in \mathbb{R}^{d_{audio}}
\tag{15}\label{eq:15}
\end{equation}

\textbf{(3) Global pooling representation:}

\begin{equation}
a = \frac{1}{M_{a}} \sum_{m = 1}^{M_{a}} u_{m} \in \mathbb{R}^{d_{audio}}
\tag{16}\label{eq:16}
\end{equation}

After type-specific encoding, the TFC inputs form a structured set of
language-like embeddings for the multimodal backbone. This serialized
representation preserves the association between fault signatures and
their physical context within one framework. Consequently, it addresses
the data-heterogeneity challenge identified in Section~\ref{sec:1} and enables
multi-object and multi-scenario diagnosis across heterogeneous machines
and multi-source datasets.

\section{Architecture of the Multimodal-Fusion Fault-Diagnosis Foundation Model}\label{sec:3}

Unlike high-density, semantically structured text in NLP, industrial
sensor signals are high-dimensional, sparse, and characterized by low
information density. A single vibration sequence may contain thousands
of data points, yet critical fault signatures are often confined to
transient impacts or weak periodic components. Directly processing the
full sequence with a generic LLM would be computationally expensive and
could obscure these features. Therefore, TFC-MMFD employs a hierarchical
architecture that progressively abstracts, refines, and fuses multimodal
inputs, which is computationally efficient and semantically effective.

\subsection{Overall architecture: hierarchical processing, feature integration, and deep fusion}\label{sec:3.1}

TFC-MMFD formulates universal fault diagnosis as semantic matching
between fault representations and natural-language descriptions rather
than classification on a fixed label space. To accommodate low-density
industrial signals, the architecture processes the inputs in stages.
Specifically, it first reorganizes the heterogeneous modalities by
physical meaning and aggregates them into compact representations. These
representations interact with fault-mode text before entering the
language-model backbone, and a set of learnable queries then extracts
mode-specific features for text-anchor matching. This
``perceive-understand-diagnose'' flow enables cross-modal
complementarity while preserving subtle condition- and object-specific
cues that distinguish similar faults.

The proposed TFC-MMFD architecture (Fig.~\ref{fig:3}) follows the principles of
``hierarchical processing, feature integration, and deep fusion'' and
comprises five core modules:

\begin{figure}[htbp]
\centering
\includegraphics[width=0.78\textwidth]{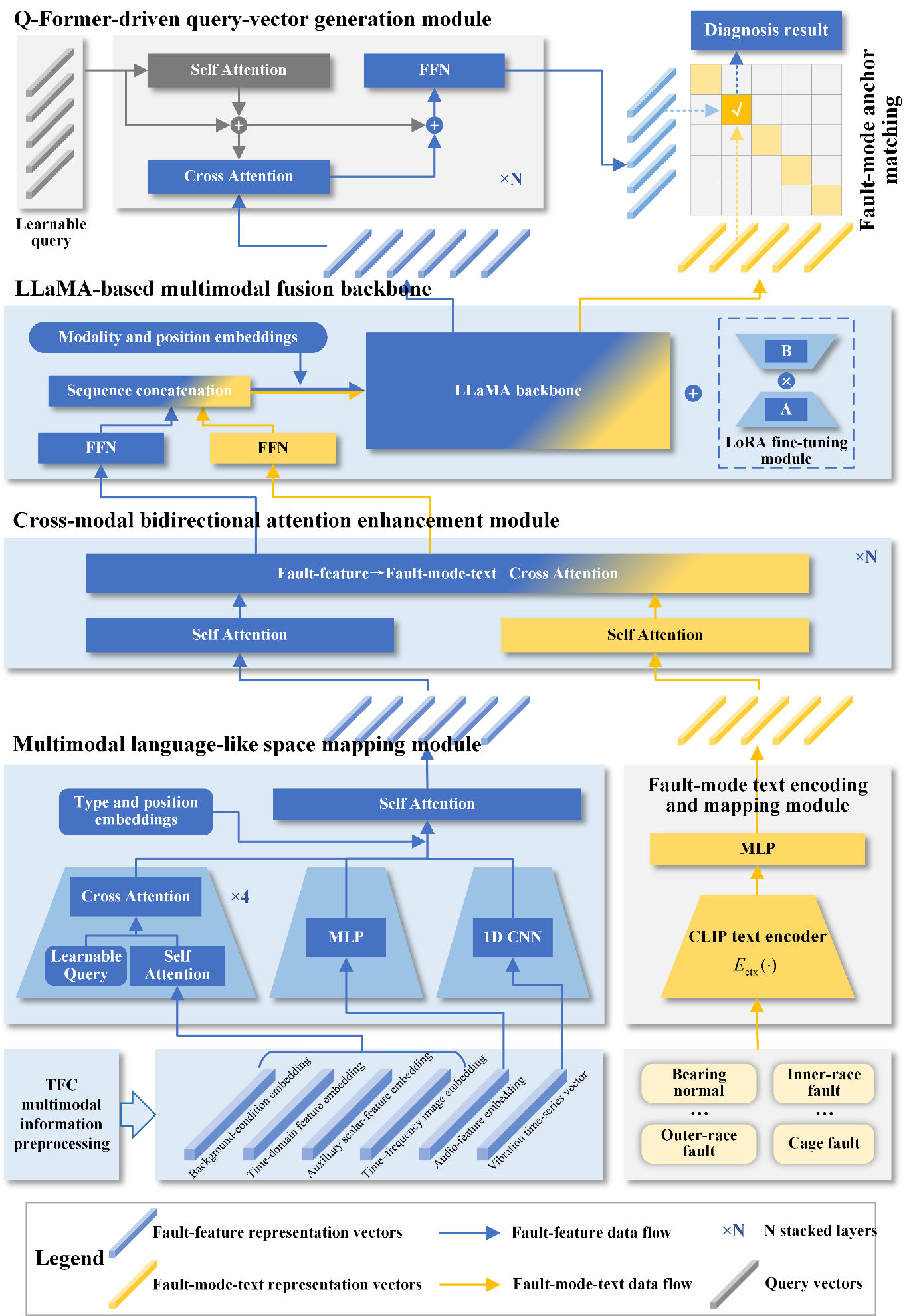}
\caption{Overall architecture of the multimodal fusion fault-diagnosis foundation model.}\label{fig:3}
\end{figure}

\begin{enumerate}
\item \textbf{Multimodal language-like space mapping module:} receives the
  tokenized TFC embeddings, re-classifies them by physical meaning, and
  maps multi-view heterogeneous features into a unified high-dimensional
  semantic space to form a standardized multimodal sequence.
\item \textbf{Fault-mode text encoding and mapping module:} converts
  fault-mode text descriptions through a pre-trained text encoder into
  vectors aligned with the multimodal feature space, thereby serving as
  semantic anchors for matching and classification.
\item \textbf{Cross-modal bidirectional attention enhancement module:}
  bridges fault features and fault-mode text through multi-layer
  cross-attention for deep interaction and alignment.
\item \textbf{LLaMA-based multimodal backbone:} feeds aligned multimodal
  features and text into the LLaMA backbone for higher-level
  representation learning, with LoRA for efficient fine-tuning.
\item \textbf{Q-Former-driven query-vector generation module:} generates a
  fixed set of learnable query vectors from the deep fault features
  distilled by LLaMA; each query captures discriminative features of a
  specific fault mode and, by similarity matching with text anchors,
  performs the final classification.
\end{enumerate}

\subsection{Multimodal language-like space mapping module: physical-meaning re-organization and query-based aggregation}\label{sec:3.2}

Although the TFC modalities have been converted into language-like
embeddings of uniform dimension, the unification remains formal: the
vectors differ in physical meaning, information density, and importance.
A dedicated mapping module re-organizes, refines, and aggregates them
into a focused, semantically concentrated high-level representation
before they enter the backbone.

\textbf{(1) Mapping-category partition.} The encoding in Section~\ref{sec:2.2} is
organized primarily by data type, but in the fault-diagnosis process
data should be analyzed according to their physical meaning rather than
their storage form: two embeddings of the same type may play entirely
different diagnostic roles (e.g., a scalar load value versus a scalar
kurtosis value). Therefore, after encoding, the embeddings are
re-organized by physical meaning into six categories:
background-condition embeddings \({Z_{i}}^{(ctx)}\), vibration
time-series vectors \({Z_{i}}^{(seq)}\), time-domain feature embeddings
\({Z_{i}}^{(time)}\), time-frequency image embeddings
\({Z_{i}}^{(img)}\), audio-feature embeddings \({Z_{i}}^{(audio)}\), and
auxiliary scalar-feature embeddings \({Z_{i}}^{(aux)}\), each expressed
in matrix form:

\begin{equation}
{Z_{i}}^{(ctx)} = E_{ctx}(\{ B_{i}, W_{i}\}) \in \mathbb{R}^{n_{ctx} \times d_{text}}
\tag{17a}\label{eq:17a}
\end{equation}

\begin{equation}
{Z_{i}}^{(seq)} = \mathrm{Prep}(x_{v,i}) \in \mathbb{R}^{1 \times d_{seq}}
\tag{17b}\label{eq:17b}
\end{equation}

\begin{equation}
{Z_{i}}^{(time)} = \Phi(\{{TD}_{i}\}) \in \mathbb{R}^{n_{time} \times d_{scalar}}
\tag{17c}\label{eq:17c}
\end{equation}

\begin{equation}
{Z_{i}}^{(img)} = E_{img}(\{ P_{i}\}) \in \mathbb{R}^{n_{img} \times d_{img}}
\tag{17d}\label{eq:17d}
\end{equation}

\begin{equation}
{Z_{i}}^{(audio)} = E_{audio}(x_{v,i}) \in \mathbb{R}^{1 \times d_{audio}}
\tag{17e}\label{eq:17e}
\end{equation}

\begin{equation}
{Z_{i}}^{(aux)} = \Phi(\{{{TD}_{i}}^{aux}, {{TF}_{i}}^{aux}, {{FD}_{i}}^{aux}\}) \in \mathbb{R}^{n_{aux} \times d_{scalar}}
\tag{17f}\label{eq:17f}
\end{equation}

\noindent where d\_seq denotes the length of the vibration sequence after the
preprocessing operation Prep(\(\cdot\)).

Rationale behind the partitioning scheme:

\begin{steps}
\item The background-condition category, synthesized from object-structure
and operating-condition information, provides machine priors and
environmental context that act as physical constraints.
\item The four core categories each characterize the vibration from a
distinct viewpoint and contribute directly to fault identification, and
are therefore kept separate.
\item Auxiliary scalar features provide only limited, local evidence and
are consolidated into one auxiliary category.
\end{steps}

\textbf{(2) Query-based attention-aggregation mapping.} For the four
multi-field categories, let intra-modal sub-vectors be
\(Z = \{ z_{1}, \ldots, z_{n}\}, z_{k} \in \mathbb{R}^{d_{k}}\).

\begin{steps}
\item Linear projection unifies the dimensions to form a sequence \(Z'\):
\begin{equation}
{z_{k}}^{'} = {Linear}_{k}(z_{k}),\quad {z_{k}}^{'} \in \mathbb{R}^{d};\quad Z' = \lbrack{z_{1}}^{'}; \ldots; {z_{n}}^{'}\rbrack \in \mathbb{R}^{n \times d}
\tag{18}\label{eq:18}
\end{equation}
\item Self-attention models intra-modal field dependencies:
\begin{equation}
Z'' = \mathrm{SelfAttn}(Z') \in \mathbb{R}^{n \times d}
\tag{19}\label{eq:19}
\end{equation}
\item Cross-attention with a learnable query \(\tilde{q}\) aggregates key information
into a single global vector:
\begin{equation}
\tilde{z} = \mathrm{CrossAttn}(\tilde{q}, Z'') \in \mathbb{R}^{d}
\tag{20}\label{eq:20}
\end{equation}
\end{steps}

\noindent yielding
\({{\tilde{z}}_{i}}^{(ctx)}, {{\tilde{z}}_{i}}^{(time)}, {{\tilde{z}}_{i}}^{(img)}, {{\tilde{z}}_{i}}^{(aux)}\).
This design adapts flexibly to heterogeneous object backgrounds across
machines without altering the backbone.

\textbf{(3) MLP single-vector projection.} The audio embedding
\({z_{i}}^{(audio)}\) is aligned via an MLP (fully-connected layers,
activation \(\sigma\), GroupNorm GN):

\begin{equation}
h_{1} = \sigma(GN({z_{i}}^{(audio)}W_{1} + b_{1}));\quad {{\tilde{z}}_{i}}^{(audio)} = \sigma(GN(h_{1}W_{2} + b_{2}))
\tag{21}\label{eq:21}
\end{equation}

\textbf{(4) 1-D CNN sequence projection.} The raw vibration time-series
\({z_{i}}^{(seq)}\) preserves the most complete information. A two-layer
Conv1d with GELU \(\varphi\), GroupNorm, and global pooling extracts and
down-samples it:

\begin{equation}
v = \mathrm{Pool}(\varphi({Conv1d}_{2}(\varphi({Conv1d}_{1}({z_{i}}^{(seq)})))));\quad {{\tilde{z}}_{i}}^{(seq)} = \mathrm{LayerNorm}(\mathrm{GELU}(vW + b)) \in \mathbb{R}^{d}
\tag{22}\label{eq:22}
\end{equation}

\textbf{(5) Type/position embedding and preliminary fusion.} The six
projected vectors are stacked into a six-element multimodal sequence
\(M_{i}\); learnable type and position embeddings are added:

\begin{equation}
{\tilde{M}}_{i} = M_{i} + \mathrm{TypeEmbedding}(type\_id) + \mathrm{PositionEmbedding}(pos\_id)
\tag{23}\label{eq:23}
\end{equation}

A multi-layer Transformer encoder then performs preliminary fusion:

\begin{equation}
F_{i} = \mathrm{TransformerEncoder}({\tilde{M}}_{i}) \in \mathbb{R}^{6 \times d}
\tag{24}\label{eq:24}
\end{equation}

\subsection{Fault-mode text encoding and mapping module: semantic fault-mode anchors}\label{sec:3.3}

Rather than mapping fault features to numerical class indices, TFC-MMFD
represents each diagnostic target as a natural-language description and
maps it to a fault-mode anchor in the shared latent space. This anchor
mechanism, a core innovation of the model, provides three key
advantages. First, it associates each diagnosis with a named fault
rather than an opaque numerical label. Second, it allows fault
descriptions to be added without retraining a fixed classification head
or reconstructing the label space. Third, it combines fault modes from
structurally different machines in one anchor library, providing a
flexible output space for different deployment settings.

Let \({T_{i}}^{(fault)}\) denote the fault-mode text of the i-th sample.
Consistent with Section~\ref{sec:2.2.2}, the CLIP text encoder maps it into the
unified language space:

\begin{equation}
{z_{i}}^{(fault)} = E_{ctx}(\mathrm{Tokenizer}({T_{i}}^{(fault)})) \in \mathbb{R}^{d_{text}}
\tag{25}\label{eq:25}
\end{equation}

An MLP layer then adjusts the dimension and smooths the features for
alignment with the other modal embeddings:

\begin{equation}
{h_{i}}^{(fault)} = \mathrm{MLP}({z_{i}}^{(fault)}) \in \mathbb{R}^{d}
\tag{26}\label{eq:26}
\end{equation}

The target fault-mode text set, the anchor library, is denoted:

\begin{equation}
T^{(fault)} = \lbrack{h_{1}}^{(fault)}; {h_{2}}^{(fault)}; \ldots; {h_{N_{t}}}^{(fault)}\rbrack \in \mathbb{R}^{N_{t} \times d}
\tag{27}\label{eq:27}
\end{equation}

\subsection{Cross-modal bidirectional attention enhancement module: feature-text deep interaction and alignment}\label{sec:3.4}

A multi-layer Transformer with bidirectional cross-modal attention
aligns fault-mode text with the fused fault-feature sequence. This
module allows subsequent diagnosis to be guided by natural-language
descriptions: fault features learn to perceive the semantic guidance
from text, while text representations focus on critical signal details.
The resulting high-quality representations provide a solid foundation
for contrastive learning and underpin the flexible, generalizable
anchor-based diagnosis mechanism.

\textbf{(1) Learnable position encoding} is added to both projection
outputs:

\begin{equation}
T^{(fault,pe)} = T^{(fault,0)} + {PosEmbed}_{txt}\lbrack:N_{t}\rbrack,\quad F^{(pe)} = F^{(0)} + {PosEmbed}_{fuse}\lbrack:N_{f}\rbrack
\tag{28}\label{eq:28}
\end{equation}

\textbf{(2) Multi-layer self- and cross-attention.} The network has
\(L_{c}\) layers. With \(T^{(fault)(0)} = T^{(fault,pe)}\),
\(F^{(0)} = F^{(pe)}\):

\begin{steps}
\item Fault-text self-attention,
\begin{equation}
T^{(fault,sa) (l)} = \mathrm{SelfAttn}(T^{(fault)(l - 1)})
\tag{29}\label{eq:29}
\end{equation}
\item Text\(\rightarrow\)feature cross-attention (text as Query, features as Key/Value),
\begin{equation}
T^{(fault,ca) (l)} = \mathrm{CrossAttn}(T^{(fault,sa) (l)}, F^{(l - 1)})
\tag{30}\label{eq:30}
\end{equation}
\item Feature self-attention,
\begin{equation}
F^{(sa)(l)} = \mathrm{SelfAttn}(F^{(l - 1)})
\tag{31}\label{eq:31}
\end{equation}
\item Feature\(\rightarrow\)text cross-attention, completing bidirectional fusion,
\begin{equation}
F^{(l)} = \mathrm{CrossAttn}(F^{(sa) (l)}, T^{(fault,ca) (l)})
\tag{32}\label{eq:32}
\end{equation}
\end{steps}

Stacking the layers provides an iterative, high-order fusion mechanism
that ensures deep alignment between the text and feature sides.

\subsection{LLaMA-based multimodal fusion backbone: high-level abstraction and reasoning with LoRA fine-tuning}\label{sec:3.5}

To further enhance the model's semantic understanding of
fault-mode text and signal-level fault features, the pre-trained LLaMA
model is introduced as the information-fusion backbone~\cite{ref26,ref27}. The
motivation is twofold: First, the strong language representation
capability of LLaMA enables deeper cross-modal interaction and alignment
than shallow fusion networks. Second, given the scarcity of industrial
fault data, we freeze the pre-trained weights and adopt Low-Rank
Adaptation (LoRA)~\cite{ref42} for parameter-efficient fine-tuning. Let D
denote the hidden dimension of LLaMA.

\textbf{(1) Serialization and modal/position embedding.} Text and
feature sequences are projected and concatenated:

\begin{equation}
{\tilde{T}}^{(fault)} = {Proj}_{txt}(T^{(fault,L_{c})}),\quad \tilde{F} = {Proj}_{fuse}(F^{(L_{c})});\quad X = \mathrm{Concat}({\tilde{T}}^{(fault)}, \tilde{F}) \in \mathbb{R}^{(N_{t} + N_{f}) \times D}
\tag{33}\label{eq:33}
\end{equation}

\noindent with learnable modality and position embeddings added by row:

\begin{equation}
X^{(final)} = X + \mathrm{ModEmbed}(mod\_id) + \mathrm{PosEmbed}(pos\_id)
\tag{34}\label{eq:34}
\end{equation}

\textbf{(2) LLaMA forward pass:}

\begin{equation}
H = \mathrm{LLaMA}(X^{(final)})
\tag{35}\label{eq:35}
\end{equation}

After processing, \(H_{txt}, H_{feature}\) are read out from the
corresponding rows.

\textbf{(3) LoRA fine-tuning.} For a frozen projection matrix W, with
rank \(r \ll \min(d_{out}, d_{in})\), a low-rank decomposition is
added:

\begin{equation}
\Delta W = AB
\tag{36}\label{eq:36}
\end{equation}

\noindent so the LoRA-inserted forward map is:

\begin{equation}
y = W'x = (W + AB)x
\tag{37}\label{eq:37}
\end{equation}

\textbf{(4) Dimension-reduction projection} maps the outputs to the
Q-Former dimension \(d_{Q}\):

\begin{equation}
F^{(LLM)} = {Proj}_{llm}(H_{feature}) \in \mathbb{R}^{N_{f} \times d_{Q}};\quad T^{(fault,LLM)} = {Proj}_{llm}(H_{txt}) \in \mathbb{R}^{N_{t} \times d_{Q}}
\tag{38}\label{eq:38}
\end{equation}

Since W is frozen, only A and B are updated, reducing the number of
trainable parameters and the memory required for gradients and optimizer
states.

\subsection{Q-Former-driven query-vector generation module: fault-feature querying for anchor matching}\label{sec:3.6}

Fault-mode classification relies on matching fault-mode text and
multimodal fused features in a unified latent space. The Q-Former~\cite{ref43} generates a set of learnable query vectors
\(Q_{0} \in \mathbb{R}^{N_{q} \times d_{Q}}\) from the aligned
feature sequence \(F^{(LLM)}\), with the number of queries equal to the
number of fault modes (\(N_{q} = N_{t}\)), such that each query is
optimized to capture features associated with one specific fault mode.
Across the \(L\) Q-Former layers, self-attention models relationships
among the queries and cross-attention retrieves relevant information
from the feature sequence. Each Q-Former layer applies:

\textbf{(1) Query self-attention:}

\begin{equation}
{Q_{l}}^{(sa)} = \mathrm{SelfAttn}(Q_{l - 1})
\tag{39}\label{eq:39}
\end{equation}

\textbf{(2) Query\(\rightarrow\)feature cross-attention (query as Q, features as
K/V):}

\begin{equation}
Q_{l} = \mathrm{CrossAttn}({Q_{l}}^{(sa)}, F^{(LLM)})
\tag{40}\label{eq:40}
\end{equation}

After L rounds, the final query vectors are:

\begin{equation}
Q_{L} \in \mathbb{R}^{N_{q} \times d_{Q}}
\tag{41}\label{eq:41}
\end{equation}

The query vectors are matched against the fault-mode text
representations by similarity, and the highest-scoring pair yields the
diagnosis.

\section{Multi-Task Combined Training Methodology of the Foundation Model}\label{sec:4}

To address the core challenges of variable conditions, heterogeneous
machines, and sparse data, a multi-task training strategy is designed at
three levels:

\textbf{(1) Regression-adversarial training:} learns
condition-invariant, more generalizable fault representations.

\textbf{(2) Masked self-supervised pre-training:} improves robustness to
incomplete or noisy inputs and promotes deep cross-modal coupling.

\textbf{(3) Contrastive learning:} aligns the feature representations
with fault-mode text anchors for classification.

\subsection{Condition-disentangled representation learning via regression-adversarial training}\label{sec:4.1}

Machine structure, type, and condition background strongly influence
vibration signals. If the model learns representations that depend too
heavily on these background differences, robustness across machines and
conditions suffers. Conventional domain-adversarial adaptation removes
such dependence with a discrete domain classifier. However, operating
conditions in universal diagnosis are often continuous and
non-enumerable (e.g., arbitrary speed-load combinations), so enumerating
discrete domain labels is impractical and a newly encountered condition
would require redefining the label set. We therefore propose
regression-adversarial training: instead of a domain classifier, a
condition-background predictor (a regressor) is placed behind a Gradient
Reversal Layer (GRL)~\cite{ref44}, and the feature extractor is trained
adversarially against it. Because the supervisory target is the
continuous condition/background embedding rather than a discrete label,
the disentanglement extends naturally to unseen conditions, an important
departure from classifier-based domain-adversarial methods.

\textbf{(1) Feature aggregation.} A small Transformer aggregator forms a
global vector g from the LLaMA feature output:

\begin{equation}
g = \mathrm{Aggregator}(H_{feature}) \in \mathbb{R}^{D}
\tag{42}\label{eq:42}
\end{equation}

\textbf{(2) Gradient reversal layer.} The GRL acts as identity in the
forward pass and multiplies the gradient by \(- \lambda\) in the
backward pass:

\begin{equation}
g^{(rev)} = \mathrm{GRL}(g)
\tag{43}\label{eq:43}
\end{equation}

\textbf{(3) Condition-background predictor.} With the concatenated
condition/background embedding \(c \in \mathbb{R}^{d_{bg}}\), the
predictor regresses it from the reversed feature:

\begin{equation}
\hat{c} = BG\_\mathrm{Predictor}(g^{(rev)}) \in \mathbb{R}^{d_{bg}}
\tag{44}\label{eq:44}
\end{equation}

\textbf{(4) Adversarial objective.} The background-regression loss is
the squared error:

\begin{equation}
\mathcal{L}_{bg} = \| \hat{c} - c\|_{2}^{2}
\tag{45}\label{eq:45}
\end{equation}

During backpropagation the gradient is sign-flipped and scaled at the
GRL, so fault features retain discriminability while being encouraged to
suppress condition-related information, yielding condition-disentangled
robust representations.

\subsection{Multimodal feature fusion via condition-conditioning and masked self-supervised pre-training}\label{sec:4.2}

Masked pre-training, that is, random masking followed by latent-space
reconstruction, has become a core self-supervised technique that
improves robustness to noise and missing data and strengthens the
model's representation of structure and semantics~\cite{ref31,ref32}. It is particularly pertinent to fault diagnosis, where sensor
readings are frequently incomplete or heavily disturbed in industrial
environments; a model pre-trained to reconstruct masked features can
still infer the original signal characteristics when faced with partial
loss or noise. However, condition-invariance (removing
background/condition differences, Section~\ref{sec:4.1}) conflicts with accurate
reconstruction (which needs that very information). A
condition-conditioned design resolves this: part of the features is
randomly masked, the background vector is explicitly fused into the
masked sequence, and a decoder reconstructs the time-domain and
time-frequency elements through self-supervision.

\textbf{(1) Random masking.} Elements of the vibration sequence
\({Z_{i}}^{(seq)}\) and time-frequency image \({Z_{i}}^{(img)}\) are set
to zero with probability \(\alpha_{mask}\), giving the masked latent
\(H_{masked}\).

\textbf{(2) Background-vector injection.} The condition/background
vector \(c_{i}\) is projected by an MLP to the backbone hidden dimension
and inserted as an extra token:

\begin{equation}
C_{i} = \mathrm{BgMLP}(c_{i}),\quad C_{i} \in \mathbb{R}^{D}
\tag{46}\label{eq:46}
\end{equation}

\noindent forming \(H_{ext}\).

\textbf{(3) Decoder reconstruction.} A small Transformer encoder
aggregates the sequence into a global vector by mean pooling:

\begin{equation}
g_{ext} = \mathrm{SeqAgg}(H_{ext})
\tag{47}\label{eq:47}
\end{equation}

\noindent and a 1-D deconvolution decoder reconstructs the masked targets:

\begin{equation}
{{\hat{Z}}_{i}}^{(seq)} = {Decoder}_{seq}(g_{ext}),\quad {{\hat{Z}}_{i}}^{(img)} = {Decoder}_{img}(g_{ext})
\tag{48}\label{eq:48}
\end{equation}

\textbf{(4) Self-supervised reconstruction loss} (mean-squared error
against the unmasked data):

\begin{equation}
\mathcal{L}_{recon} = \sum_{m \in \{ seq,img\}} \|{{\hat{Z}}_{i}}^{(m)} - {Z_{i}}^{(m)}\|_{2}^{2}
\tag{49}\label{eq:49}
\end{equation}

The pre-training stage jointly minimizes the adversarial background
regression and masked reconstruction:

\begin{equation}
\mathcal{L}_{pretrain} = \alpha_{bg}\mathcal{L}_{bg} + \alpha_{recon}\mathcal{L}_{recon}
\tag{50}\label{eq:50}
\end{equation}

\noindent where \(\alpha_{bg}, \alpha_{recon}\) balance the two losses, jointly
driving the network to learn condition-insensitive features and support
reconstruction from incomplete inputs.

\subsection{Fault-mode-anchor matching diagnosis via contrastive-learning-based training}\label{sec:4.3}

During training, fault-mode texts serve as anchors in the latent space,
and diagnosis is performed by anchor-feature similarity matching.

\textbf{(1) Anchor-feature pairs.} The fused TFC information yields
\(N_{q}\) query vectors,

\begin{equation}
Q_{L} = \lbrack q_{1}, q_{2}, \ldots, q_{N_{q}}\rbrack \in \mathbb{R}^{N_{q} \times d_{Q}}
\tag{51}\label{eq:51}
\end{equation}

\noindent and the fault-mode texts yield the text anchors,

\begin{equation}
T^{(fault,LLM)} = \lbrack t_{1}, t_{2}, \ldots, t_{N_{q}}\rbrack \in \mathbb{R}^{N_{q} \times d_{Q}}
\tag{52}\label{eq:52}
\end{equation}

\noindent paired by index to form \(N_{q}\) anchor-feature pairs.

\textbf{(2) Text-anchor generation by supervised contrastive learning.}
Different fault-mode text representations should be well separated;
same-mode samples are positives and different-mode samples are negatives~\cite{ref45}:

\begin{equation}
\mathcal{L}_{cont} = \mathrm{ContrastiveLoss}(t_{1}, t_{2}, \ldots, t_{N_{q}})
\tag{53}\label{eq:53}
\end{equation}

\textbf{(3) Anchor-matching classification.} Features and anchors are
normalized and their similarity computed:

\begin{equation}
S_{qt\mathcal{,l}} = sim(q_{\mathcal{l}}, t_{\mathcal{l}}\mathcal{),\quad l = }1, \ldots, N_{q}
\tag{54}\label{eq:54}
\end{equation}

The \(N_{q}\) similarity values act as logits, and cross-entropy is
applied against the true label:

\begin{equation}
\mathcal{L}_{cls} = \mathrm{CrossEntropy}(sim(Q_{L}\lbrack i\rbrack, T^{(fault,LLM)}), \mathcal{l}_{i})
\tag{55}\label{eq:55}
\end{equation}

\textbf{(4) Total objective.} Text-anchor contrast and fault-mode
classification are jointly trained:

\begin{equation}
\mathcal{L}_{total} = \alpha_{contrast}\mathcal{L}_{cont} + \alpha_{cls}\mathcal{L}_{cls}
\tag{56}\label{eq:56}
\end{equation}

At inference, the similarity vector between query vectors and text
anchors is computed, and the index of the highest similarity gives the
predicted fault mode. A key advantage is text-side extensibility:
introducing a new fault mode only requires adding the corresponding
text, which the model maps to a new anchor, and no large-scale
label-system reconstruction is needed. For industrial scenarios with
labeled fault-mode text, a text-anchor library enables online
similarity-based prediction with flexible extensibility and semantic
interpretability.

The three training mechanisms support the capabilities evaluated in
Section~\ref{sec:5}. Regression-adversarial condition disentanglement targets
zero-shot generalization to unseen operating conditions.
Condition-conditioned masked pre-training supports reconstruction from
noisy or incomplete monitoring data and promotes cross-modal coupling.
Contrastive fault-mode-anchor matching provides discrimination among
known faults and an extensible output space for few-shot adaptation to
new machines. Together with the unified representation in Section~\ref{sec:2} and
the hierarchical architecture in Section~\ref{sec:3}, these mechanisms form the
complete TFC-MMFD pipeline.

\section{Case Study}\label{sec:5}

Industrial fault-diagnosis models must accommodate operating-condition
changes caused by variations in speed and load, as well as distribution
shifts when a model is transferred to an unseen machine. This section
evaluates the performance, generalization, and adaptability of TFC-MMFD
through four tasks in three groups, progressing from in-distribution
diagnosis to cross-machine adaptation:

\begin{enumerate}
\item \textbf{Base universal diagnosis capability (Task 1).} Under the
  relatively ideal condition that training and test data share a
  consistent distribution, this tests whether the model can unify ten
  multi-source datasets into one model and achieve high-accuracy
  diagnosis, establishing a performance baseline.
\item \textbf{Cross-condition generalization capability (Task 2).} This
  tests model behavior on operating conditions never seen during
  training, that is, zero-shot generalization without any additional
  training. This task probes the condition-disentanglement ability of
  the model.
\item \textbf{Cross-machine (cross-dataset) generalization and adaptation
  capability (Tasks 3 and 4).} This tests zero-shot generalization to
  entirely new machine objects (Task 3) and the performance after
  fine-tuning with only a handful of target-domain samples (Task 4),
  reflecting the model's value as a pre-trained
  foundation model.
\end{enumerate}

Each task below is described by its objective and its experimental
setup, followed by a detailed result analysis. Sections 5.7 and 5.8 then
provide a comparative analysis against representative baselines and an
overall summary.

\subsection{Heterogeneous benchmark of ten public bearing datasets}\label{sec:5.1}

\begin{figure}[htbp]
\centering
\begin{minipage}{0.48\textwidth}\centering
\includegraphics[width=\linewidth]{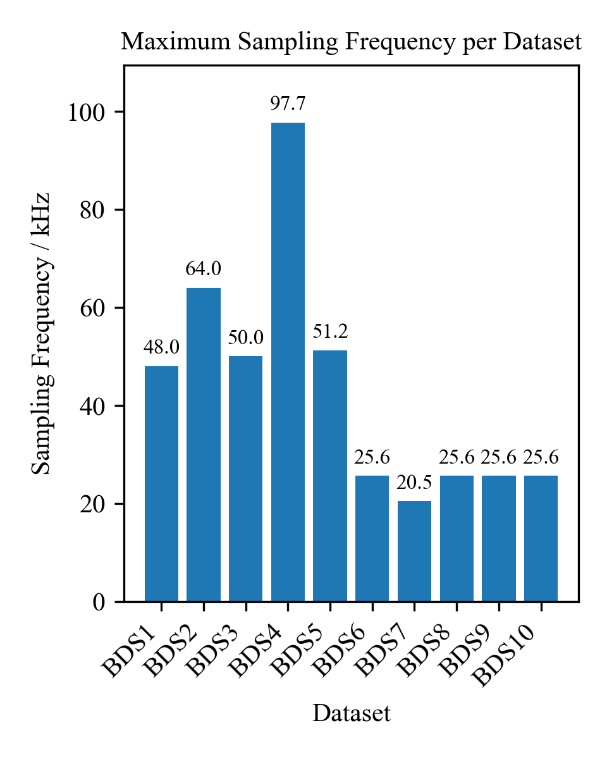}
\end{minipage}
\hfill
\begin{minipage}{0.48\textwidth}\centering
\includegraphics[width=\linewidth]{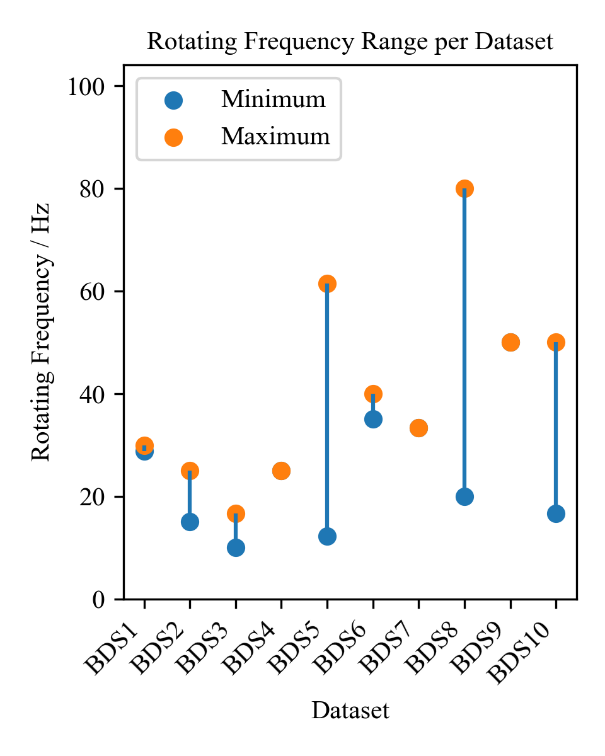}
\end{minipage}
\caption{Statistics of sampling frequency and rotational-frequency range across datasets.}\label{fig:4}
\end{figure}

Ten publicly available bearing fault datasets are used: CWRU~\cite{ref46},
PU~\cite{ref47}, JNU~\cite{ref48}, MFPT~\cite{ref49}, MAFAULDA~\cite{ref50}, XJTU~\cite{ref51}, IMS~\cite{ref52}, HUST~\cite{ref13}, SQV~\cite{ref53,ref54}, and SDUST~\cite{ref55,ref56}. The datasets differ in source, machine type, condition
coverage, and fault modes, providing a heterogeneous benchmark for
evaluating TFC-MMFD. A summary is given in Table~\ref{tab:3}, and the
sampling-frequency and rotational-frequency statistics are visualized in
Fig.~\ref{fig:4}.

\begin{longtable}[]{@{}
  >{\raggedright\arraybackslash}p{(\columnwidth - 22\tabcolsep) * \real{0.0718}}
  >{\raggedright\arraybackslash}p{(\columnwidth - 22\tabcolsep) * \real{0.0797}}
  >{\raggedright\arraybackslash}p{(\columnwidth - 22\tabcolsep) * \real{0.0757}}
  >{\raggedright\arraybackslash}p{(\columnwidth - 22\tabcolsep) * \real{0.0908}}
  >{\raggedright\arraybackslash}p{(\columnwidth - 22\tabcolsep) * \real{0.0757}}
  >{\raggedright\arraybackslash}p{(\columnwidth - 22\tabcolsep) * \real{0.0908}}
  >{\raggedright\arraybackslash}p{(\columnwidth - 22\tabcolsep) * \real{0.0795}}
  >{\raggedright\arraybackslash}p{(\columnwidth - 22\tabcolsep) * \real{0.0765}}
  >{\raggedright\arraybackslash}p{(\columnwidth - 22\tabcolsep) * \real{0.0880}}
  >{\raggedright\arraybackslash}p{(\columnwidth - 22\tabcolsep) * \real{0.0812}}
  >{\raggedright\arraybackslash}p{(\columnwidth - 22\tabcolsep) * \real{0.0908}}
  >{\raggedright\arraybackslash}p{(\columnwidth - 22\tabcolsep) * \real{0.0994}}@{}}
\caption{Overview of the bearing datasets used in the case study.}\label{tab:3}\\
\toprule\noalign{}
\begin{minipage}[b]{\linewidth}\raggedright
\textbf{ID}
\end{minipage} & \begin{minipage}[b]{\linewidth}\raggedright
\textbf{Dataset}
\end{minipage} & \begin{minipage}[b]{\linewidth}\raggedright
\textbf{Object}
\end{minipage} & \begin{minipage}[b]{\linewidth}\raggedright
\textbf{RF code}
\end{minipage} & \begin{minipage}[b]{\linewidth}\raggedright
\textbf{RF (Hz)}
\end{minipage} & \begin{minipage}[b]{\linewidth}\raggedright
\textbf{LD code}
\end{minipage} & \begin{minipage}[b]{\linewidth}\raggedright
\textbf{Load}
\end{minipage} & \begin{minipage}[b]{\linewidth}\raggedright
\textbf{Cond. combos}
\end{minipage} & \begin{minipage}[b]{\linewidth}\raggedright
\textbf{Fault modes}
\end{minipage} & \begin{minipage}[b]{\linewidth}\raggedright
\textbf{Severity levels}
\end{minipage} & \begin{minipage}[b]{\linewidth}\raggedright
\textbf{Sampling (Hz)}
\end{minipage} & \begin{minipage}[b]{\linewidth}\raggedright
\textbf{Bearing type}
\end{minipage} \\
\midrule\noalign{}
\endhead
\bottomrule\noalign{}
\endlastfoot
BDS1 & CWRU & Bearing & RF1\_1

RF1\_2

RF1\_3

RF1\_4 & 29.95

29.53

29.17

28.83 & LD1\_1

LD1\_2

LD1\_3

LD1\_4 & 0 HP

1 HP

2 HP

3 HP & 4 & Normal

IR

OR

Ball & 4 & 12k / 48k & Deep-groove ball \\
\midrule
BDS2 & PU & Bearing & RF2\_1

RF2\_2 & 15

25 & LD2\_1

LD2\_2 & 0.1 Nm

0.7 Nm & 3 & Normal

IR

OR & 2 & 64k & Deep-groove ball \\
\midrule
BDS3 & JNU & Bearing & RF3\_1

RF3\_2

RF3\_3 & 10

13.33

16.67 & --- & --- & 3 & Normal

IR

OR

Ball & 2 & 50k & Cylindrical roller \\
\midrule
BDS4 & MFPT & Bearing & RF4\_1 & 25 & LD4\_1

LD4\_2

LD4\_3

LD4\_4

LD4\_5

LD4\_6

LD4\_7

LD4\_8

LD4\_9 & 0 lbs

25 lbs

50 lbs

100 lbs

150 lbs

200 lbs

250 lbs

270 lbs

300 lbs & 9 & Normal

IR

OR & 1 & 48.8k / 97.7k & Deep-groove ball \\
\midrule
BDS5 & MAFAULDA & Bearing & RF5\_1-

RF5\_51 & 12.28-

61.43 & LD5\_1-

LD5\_43 & --- & 1951 & Normal

OR

Ball

Cage & 1 & 51.2k & Insert ball \\
\midrule
BDS6 & XJTU & Bearing & RF6\_1

RF6\_2

RF6\_3 & 35

37.5

40 & LD6\_1

LD6\_2

LD6\_3 & 10 kN

11 kN

12 kN & 3 & Normal

IR

OR

Cage & 1 & 25.6k & Insert ball \\
\midrule
BDS7 & IMS & Bearing & RF7\_1 & 33.33 & LD7\_1 & 6000 lbs & 1 & Normal

IR

OR

Ball & 1 & 20.48k & Double-row roller \\
\midrule
BDS8 & HUST & Bearing & RF8\_1-

RF8\_10 & 20

25

30

35

40

60

65

70

75

80 & --- & --- & 10 & Normal

IR

OR

Ball & 2 & 25.6k & Insert ball \\
\midrule
BDS9 & SQV & Bearing & RF9\_1 & 50 & --- & --- & 1 & Normal

IR

OR & 3 & 25.6k & Deep-groove ball \\
\midrule
BDS10 & SDUST & Bearing & RF10\_1

RF10\_2

RF10\_3

RF10\_4

RF10\_5

RF10\_6 & 16.67

25

30

33.33

41.67

50 & LD10\_1

LD10\_2

LD10\_3

LD10\_4 & 0 N

20 N

40 N

60 N & 24 & Normal

IR

OR

Ball & 3 & 25.6k & Deep-groove ball \\
\end{longtable}

\emph{RF: rotational-frequency code; LD: load code; IR: inner-race
fault; OR: outer-race fault; Ball: rolling-element fault; Cage: cage
fault.}

The datasets differ in four main respects:

\begin{enumerate}
\item \textbf{Bearing type:} deep-groove ball, cylindrical roller, and
  insert-type spherical bearings, with different physical structures and
  vibration-transmission paths.
\item \textbf{Operating conditions:} speeds from 10 Hz to 80 Hz and loads
  from light to heavy, forming many condition combinations.
\item \textbf{Fault modes:} inner-race, outer-race, rolling-element, and
  cage faults at several severity levels.
\item \textbf{Data acquisition:} sampling frequency spanning 12 kHz to
  97.656 kHz, posing a challenge for unified modeling.
\end{enumerate}

\subsection{Model and training settings}\label{sec:5.2}

All experiments use the architecture of Section~\ref{sec:3}. The key component
settings are organized below by module, using the same module names as
in the architecture and training sections; full parameters are listed in
Tables~\ref{tab:4} and~\ref{tab:5}.

\begin{enumerate}
\item \textbf{Multimodal language-like space mapping module (Section~\ref{sec:3.2}):}
  all input modalities (time-domain, frequency-domain, and condition)
  are first projected to a unified 1024-dimensional token space; the
  tokenized feature sequence is then fed to a fusion module of three
  Transformer encoder layers with 8-head multi-head self-attention for
  deep interaction and aggregation.
\item \textbf{Cross-modal bidirectional attention enhancement module
  (Section~\ref{sec:3.4}):} three stacked cross-attention layers with 8 heads and
  a hidden dimension of 1024 perform feature-text deep interaction and
  alignment.
\item \textbf{LLaMA-based multimodal fusion backbone with LoRA (Section~\ref{sec:3.5}):} LLaMA-3.2-3B (hidden dimension 3072) serves as the
  knowledge-encoding and reasoning core; rank-8 LoRA (scaling factor 32,
  dropout 0.05) is applied to the query (q\_proj) and value (v\_proj)
  projection matrices of each Transformer layer.
\item \textbf{Q-Former-driven query-vector generation module (Section~\ref{sec:3.6}):}
  a two-layer Q-Former with 5 learnable query vectors, equal to the
  number of core fault categories (normal plus four faults), guides each
  query to focus on the abstract representation of a specific fault
  mode.
\item \textbf{Pre-training task (Section~\ref{sec:4.2}):} 15\% of the input features
  are randomly masked to drive the adversarial background-prediction and
  multimodal-reconstruction objectives.
\end{enumerate}

\begin{longtable}[]{@{}
  >{\raggedright\arraybackslash}p{(\columnwidth - 6\tabcolsep) * \real{0.2121}}
  >{\raggedright\arraybackslash}p{(\columnwidth - 6\tabcolsep) * \real{0.2271}}
  >{\raggedright\arraybackslash}p{(\columnwidth - 6\tabcolsep) * \real{0.1817}}
  >{\raggedright\arraybackslash}p{(\columnwidth - 6\tabcolsep) * \real{0.3791}}@{}}
\caption{Key component parameters of the fault-diagnosis foundation model.}\label{tab:4}\\
\toprule\noalign{}
\begin{minipage}[b]{\linewidth}\raggedright
\textbf{Module / Component}
\end{minipage} & \begin{minipage}[b]{\linewidth}\raggedright
\textbf{Parameter}
\end{minipage} & \begin{minipage}[b]{\linewidth}\raggedright
\textbf{Value}
\end{minipage} & \begin{minipage}[b]{\linewidth}\raggedright
\textbf{Description}
\end{minipage} \\
\midrule\noalign{}
\endhead
\bottomrule\noalign{}
\endlastfoot
\multirow{3}{=}{Multimodal mapping} & Target token dim. & 1024 & Unified
projection dim. of all modalities. \\*
& Fusion Transformer layers & 3 & Encoder layers fusing multimodal
tokens. \\*
& Fusion attention heads & 8 & Heads of the multi-head
self-attention. \\
\midrule
\multirow{3}{=}{Cross-modal attention} & Hidden dim. & 1024 & Internal
dim. of text-feature cross-attention. \\*
& Cross-attention layers & 3 & Number of stacked cross-attention
layers. \\*
& Cross-attention heads & 8 & Heads of multi-head cross-attention. \\
\midrule
\multirow{2}{=}{LLaMA backbone} & Model & LLaMA-3.2-3B & Pre-trained
language foundation model. \\*
& Hidden dim. & 3072 & Internal hidden dimension of LLaMA. \\
\midrule
\multirow{4}{=}{LoRA fine-tuning} & Rank & 8 & Rank of the low-rank
matrices. \\*
& Scaling factor & 32 & LoRA scaling coefficient. \\*
& Dropout & 0.05 & Dropout applied to LoRA modules. \\*
& Target modules & q\_proj, v\_proj & Q and V projection matrices in
LLaMA. \\
\midrule
\multirow{3}{=}{Q-Former} & Query vectors & 5 & Equal to the number of
fault categories. \\*
& Hidden dim. & 1024 & Internal Transformer hidden dim. \\*
& Layers & 2 & Number of Transformer blocks. \\
\end{longtable}

\begin{longtable}[]{@{}
  >{\raggedright\arraybackslash}p{(\columnwidth - 2\tabcolsep) * \real{0.5680}}
  >{\raggedright\arraybackslash}p{(\columnwidth - 2\tabcolsep) * \real{0.4320}}@{}}
\caption{Core training parameters.}\label{tab:5}\\
\toprule\noalign{}
\begin{minipage}[b]{\linewidth}\raggedright
\textbf{Parameter}
\end{minipage} & \begin{minipage}[b]{\linewidth}\raggedright
\textbf{Value}
\end{minipage} \\
\midrule\noalign{}
\endhead
\bottomrule\noalign{}
\endlastfoot
Optimizer & Adam \\*
Batch size & 32 \\*
Learning rate & \(1\times 10^{-5}\) \\*
Pre-training mask ratio & 0.15 \\
\end{longtable}

\subsection{Task 1: Base universal diagnosis}\label{sec:5.3}

\textbf{(1) Task objective.} Verify that, under the ideal condition of
sufficient and distribution-consistent data, TFC-MMFD can effectively
integrate the multimodal information of ten independent datasets to
achieve high-accuracy universal fault diagnosis, thereby establishing a
performance baseline.

\textbf{(2) Task setup.} All ten bearing datasets (BDS1-BDS10) are
merged into one large mixed dataset, which is then randomly split 8:1:1
into training, validation, and test sets.

\begin{figure}[htbp]
\centering
\includegraphics[width=0.89\textwidth]{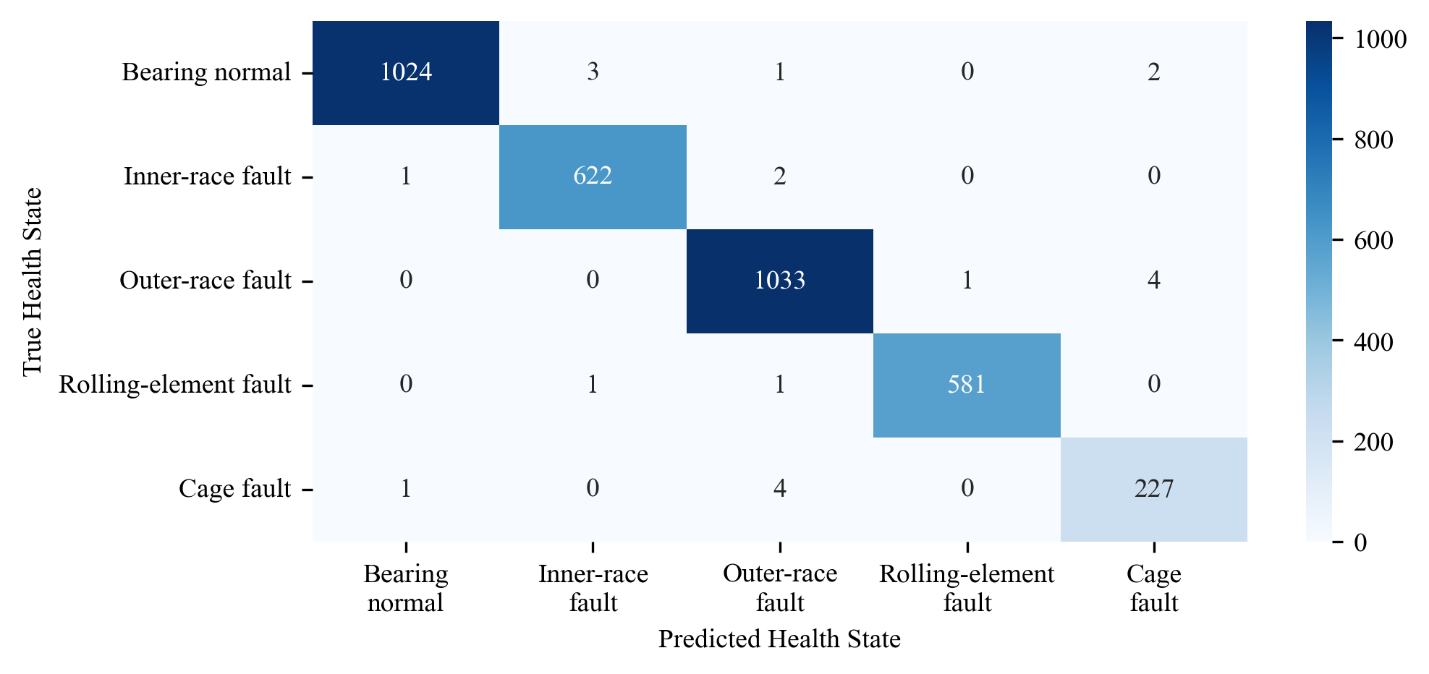}
\caption{Confusion matrix of the test set for the base universal diagnosis task.}\label{fig:5}
\end{figure}

\begin{figure}[htbp]
\centering
\includegraphics[width=0.97\textwidth]{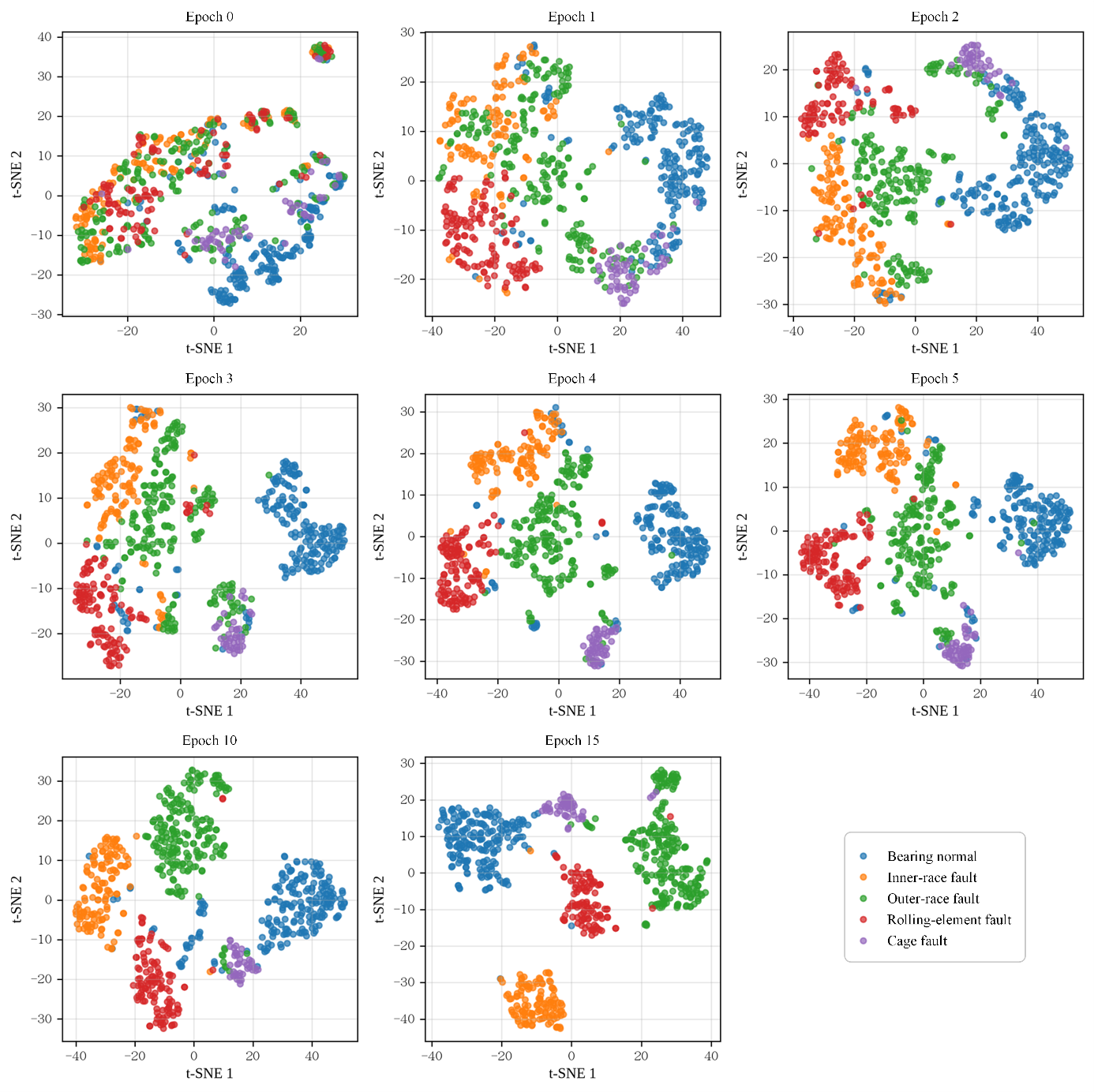}
\caption{Temporal t-SNE evolution of internal features during training.}\label{fig:6}
\end{figure}

\begin{figure}[htbp]
\centering
\includegraphics[width=1.00\textwidth]{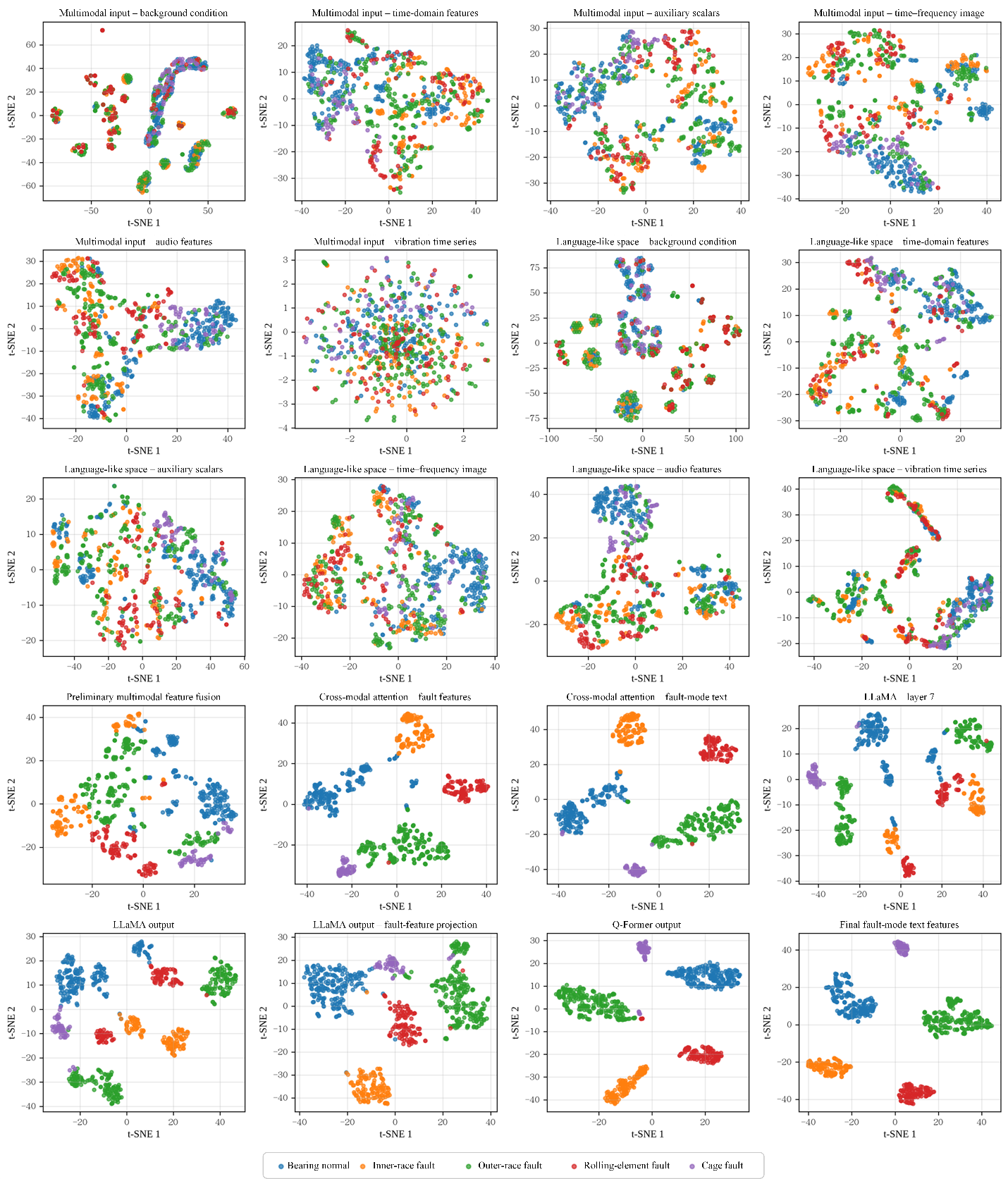}
\caption{Spatial t-SNE evolution of features from shallow to deep layers.}\label{fig:7}
\end{figure}

\textbf{(3) Result analysis.} The Task-1 results are given in Tables~\ref{tab:6}--\ref{tab:7}
and Figs.~\ref{fig:5}--\ref{fig:7}. The main observations are:

\begin{steps}
\item Overall diagnosis performance. On the test set, the model attains
99.40\% overall accuracy, a 99.16\% macro-average F1, a 99.40\%
weighted-average F1, and 99.77\% accuracy on binary anomaly detection
(Table~\ref{tab:6}). The main observations are:
\begin{steps}
\item Universality: under ten heterogeneous-object datasets and
  differentiated operating conditions, the 99.40\% overall accuracy,
  with no systematic confusion across objects or conditions,
  demonstrates the effectiveness of the architecture. Converting time-,
  frequency-, and condition-domain information into language-like tokens
  and processing them with a Transformer fusion module and a LLaMA
  backbone enables the model to learn stable, discriminative fault
  features from heterogeneous, multi-source data, which is the core
  capability of a universal diagnosis foundation model.
\item Imbalanced data: the macro-F1 (99.16\%) is very close to the
  weighted-F1 (99.40\%), showing that the model does not become biased
  toward majority fault modes. Given that the class distribution across
  the ten datasets is highly imbalanced (e.g., cage-fault samples are
  scarce), this indicates the model does not sacrifice minority-class
  recognition.
\item Anomaly detection: on the simpler binary anomaly-detection task the
  accuracy reaches 99.77\%, indicating that the model establishes a
  strong baseline and is highly sensitive to deviations from normal
  operation.
\end{steps}
\end{steps}

\begin{longtable}[]{@{}
  >{\raggedright\arraybackslash}p{(\columnwidth - 2\tabcolsep) * \real{0.5585}}
  >{\raggedright\arraybackslash}p{(\columnwidth - 2\tabcolsep) * \real{0.4415}}@{}}
\caption{Performance metrics for the base universal diagnosis task.}\label{tab:6}\\
\toprule\noalign{}
\begin{minipage}[b]{\linewidth}\raggedright
\textbf{Metric}
\end{minipage} & \begin{minipage}[b]{\linewidth}\raggedright
\textbf{Value}
\end{minipage} \\
\midrule\noalign{}
\endhead
\bottomrule\noalign{}
\endlastfoot
Overall accuracy & 99.40\% \\*
Macro-average F1 & 99.16\% \\*
Weighted-average F1 & 99.40\% \\*
Binary anomaly-detection accuracy & 99.77\% \\
\end{longtable}

\begin{steps}[start=2]
\item Per-class performance and confusion-matrix analysis. To probe the
model's behavior in detail, the test-set confusion
matrix (Fig.~\ref{fig:5}) and per-class metrics (Table~\ref{tab:7}) are computed. The vast
majority of samples are correctly classified; precision, recall, and F1
exceed 0.974 for all classes, with most above 0.99, again confirming the
universal diagnosis performance. The cage-fault precision (0.9742) is
slightly lower than the others: from the confusion matrix, four
outer-race and two normal samples are misclassified as cage faults,
while four cage-fault samples are misidentified as outer-race faults,
indicating a slight confusion between cage and outer-race faults in the
feature space. The likely cause is that cage-fault samples are very
scarce across the ten datasets, far fewer than the other classes, so the
model struggles to learn robust, generalizable features for this mode,
especially when its periodic impacts resemble those of outer-race faults
or background noise.
\end{steps}

\begin{longtable}[]{@{}
  >{\raggedright\arraybackslash}p{(\columnwidth - 8\tabcolsep) * \real{0.2949}}
  >{\raggedright\arraybackslash}p{(\columnwidth - 8\tabcolsep) * \real{0.1763}}
  >{\raggedright\arraybackslash}p{(\columnwidth - 8\tabcolsep) * \real{0.1763}}
  >{\raggedright\arraybackslash}p{(\columnwidth - 8\tabcolsep) * \real{0.1763}}
  >{\raggedright\arraybackslash}p{(\columnwidth - 8\tabcolsep) * \real{0.1763}}@{}}
\caption{Per-fault-mode performance for the base universal diagnosis task.}\label{tab:7}\\
\toprule\noalign{}
\begin{minipage}[b]{\linewidth}\raggedright
\textbf{Class}
\end{minipage} & \begin{minipage}[b]{\linewidth}\raggedright
\textbf{Precision}
\end{minipage} & \begin{minipage}[b]{\linewidth}\raggedright
\textbf{Recall}
\end{minipage} & \begin{minipage}[b]{\linewidth}\raggedright
\textbf{F1-Score}
\end{minipage} & \begin{minipage}[b]{\linewidth}\raggedright
\textbf{Support}
\end{minipage} \\
\midrule\noalign{}
\endhead
\bottomrule\noalign{}
\endlastfoot
Bearing normal & 0.9981 & 0.9942 & 0.9961 & 1030 \\*
Inner-race fault & 0.9936 & 0.9952 & 0.9944 & 625 \\*
Outer-race fault & 0.9923 & 0.9952 & 0.9937 & 1038 \\*
Rolling-element fault & 0.9983 & 0.9966 & 0.9974 & 583 \\*
Cage fault & 0.9742 & 0.9784 & 0.9763 & 232 \\
\end{longtable}

\begin{steps}[start=3]
\item Feature-space visualization analysis. Using t-SNE dimensionality
reduction, the evolution of the model's internal
features is analyzed along the temporal and spatial dimensions.
Temporally (Fig.~\ref{fig:6}), training exhibits a progressive chaos-to-order
feature-learning process: in the early stage (epochs 0-3) the fault
classes are highly overlapped in the t-SNE space; as training proceeds
(epochs 5-10) samples of the same class gradually gather into loose
clusters and inter-class boundaries begin to appear; in the late stage
(epoch 15) the fault types are fully separated into tight, clearly
bounded clusters, with markedly higher intra-class compactness and
inter-class separation, confirming the model's
feature-learning ability. Spatially (Fig.~\ref{fig:7}), the model constructs a
complete deep-feature pipeline from shallow to deep layers, achieving a
hierarchical mapping from low-information-density physical signals to
high-level abstract semantics: the raw-feature layer (six modalities)
has low separability; after the projection layer unifies them into the
language-like space, separability begins to improve; the multimodal
fusion layer integrates the information and markedly improves
separability; the cross-attention layer further strengthens
discrimination through cross-modal alignment; the deep LLaMA layers
provide highly abstract semantic understanding; and finally the Q-Former
and classification layer reach the best separation. Overall, the
visualization shows progressively more discriminative representations
across the network stages. Notably, the cage-fault cluster neighbors the
outer-race cluster with a few boundary points, consistent with the
slight confusion observed in the confusion matrix.
\end{steps}

In summary, the base universal diagnosis results demonstrate the
effectiveness of the overall TFC multimodal-fusion framework, especially
for heterogeneous objects and heterogeneous data. The model supports
unified multimodal modeling and extracts the shared and object-specific
knowledge required for fault diagnosis.

\subsection{Task 2: Cross-condition zero-shot generalization}\label{sec:5.4}

\textbf{(1) Task objective.} Evaluate whether a model trained on source
conditions can diagnose previously unseen operating conditions without
additional fine-tuning.

\textbf{(2) Task setup.} A leave-one-condition-out scheme is adopted
across the seven datasets that contain multiple differentiated operating
conditions. In turn, all samples of one specific condition (uniquely
determined by speed RF and load LD) form an independent unknown target
test set, while the training set comprises all available data except
that target condition. For comparison, ``zero-shot adaptation''
additionally fine-tunes the pre-trained model on the
non-target-condition samples of the target condition's
own dataset.

\begin{figure}[htbp]
\centering
\includegraphics[width=0.75\textwidth]{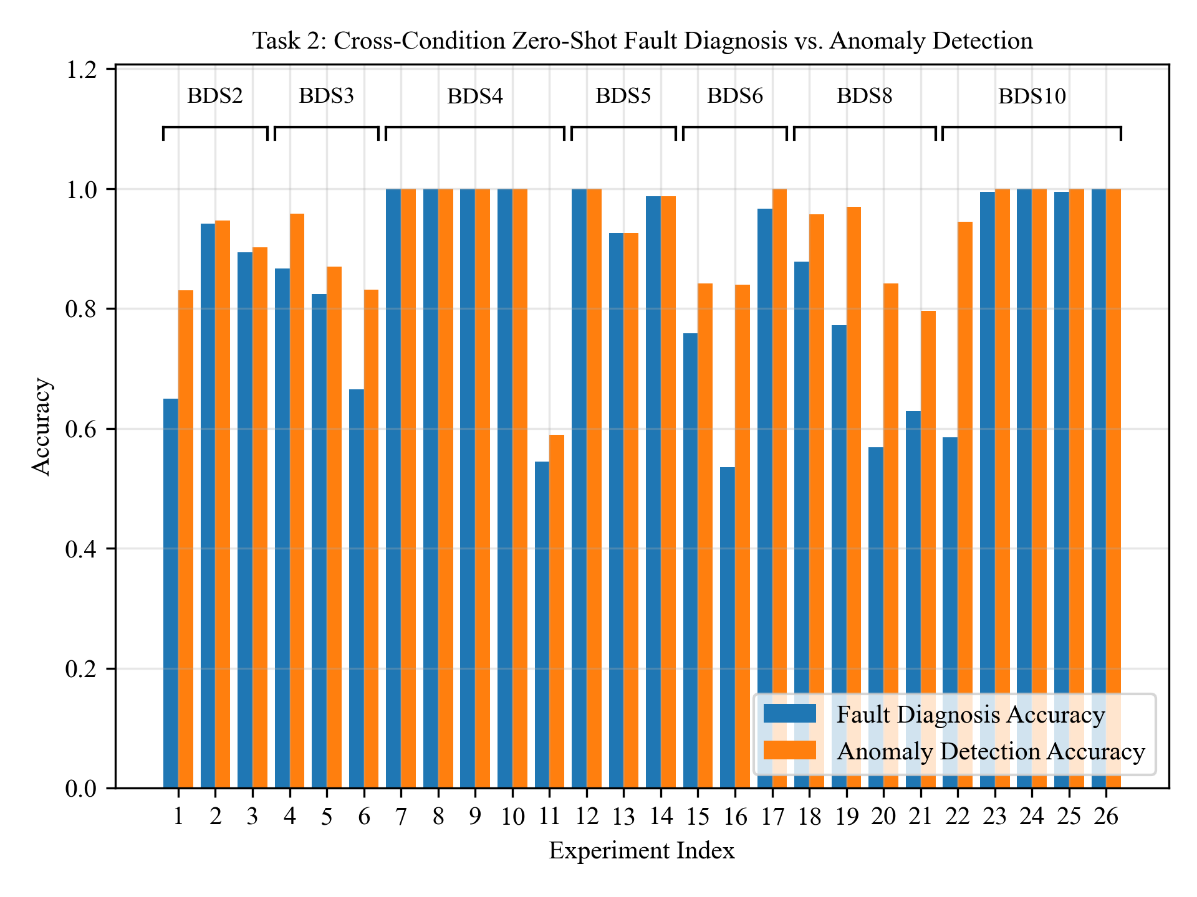}
\caption{Comparison of cross-condition zero-shot accuracy.}\label{fig:8}
\end{figure}

\begin{figure}[htbp]
\centering
\includegraphics[width=0.72\textwidth]{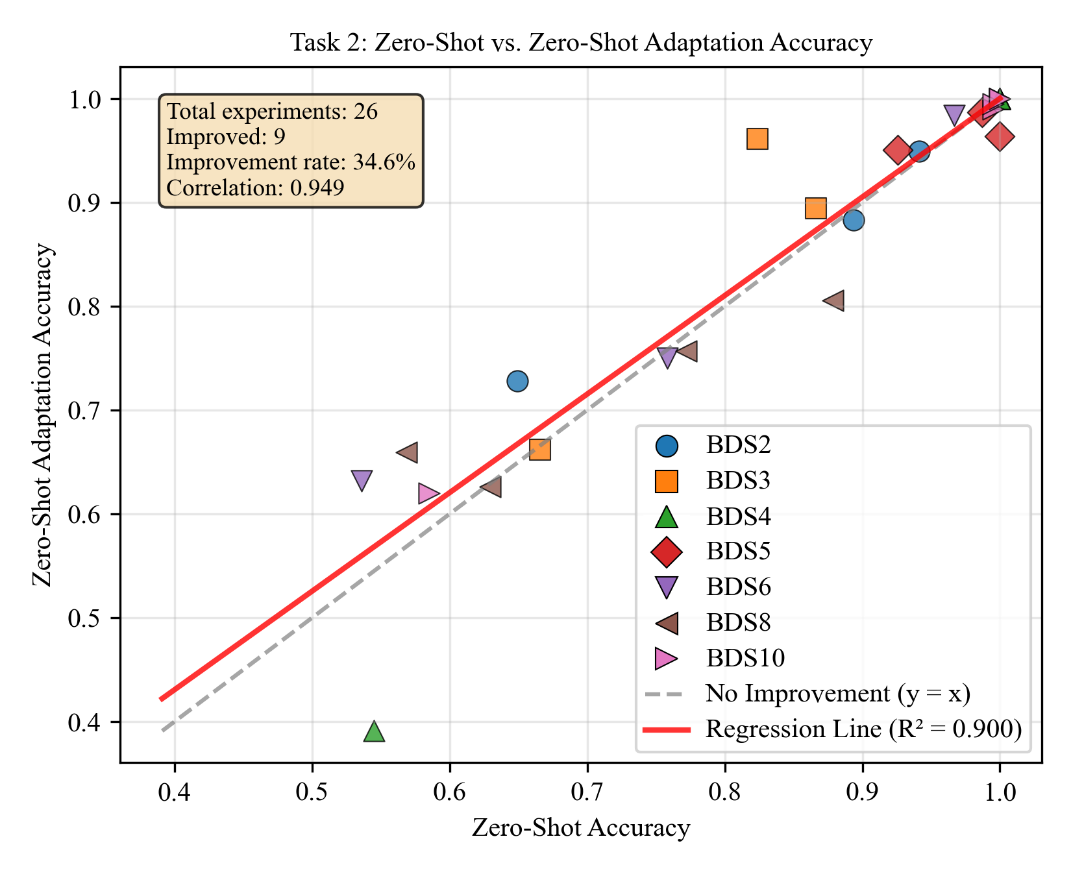}
\caption{Generalization accuracy under zero-shot vs. zero-shot adaptation.}\label{fig:9}
\end{figure}

\textbf{(3) Result analysis.} The Task-2 results are given in Table~\ref{tab:8}
and Figs.~\ref{fig:8}--\ref{fig:9}. The main observations are:

\begin{steps}
\item Relatively general cross-condition generalization: in most
  experiments, the zero-shot generalization accuracy is significantly
  higher than the random-guess level (20\%), staying above 80\% in most
  cases. When the target condition differs little from the training
  conditions, especially when it is an interpolated condition, the model
  exhibits excellent generalization. This indicates that by fusing
  condition information as a modality and performing adversarial
  training, the model has, to some degree, learned to decouple fault
  features from the influence of operating conditions.
\item Robustness to training-sample order: Fig.~\ref{fig:9} shows that the zero-shot
  and zero-shot-adaptation accuracies exhibit no significant difference,
  meaning the model is robust to the order of training samples and that
  heterogeneous samples from different datasets do not noticeably
  interfere with training. In practice, the model therefore requires no
  additional adaptation on the pre-training samples of a target dataset.
  Moreover, analysis of boundary conditions reveals that fine-tuning
  adaptation on a target dataset under boundary conditions may even
  introduce a risk of overfitting.
\item Boundary of generalization performance: the results indicate the
  boundary of zero-shot generalization. When the target condition lies
  at or beyond the edge of the training-condition range (i.e., an
  extrapolated condition), performance drops markedly. For HUST (BDS8),
  accuracy remains high (87.84\%) when the target is the lowest speed
  (RF = 20 Hz) but falls to 62.92\% at the highest speed (RF = 65 Hz);
  similar effects appear at the highest-load condition of MFPT (BDS4,
  LD4\_8) and the highest-speed condition of JNU (BDS3, RF3\_3).
\item Likely cause of extrapolation degradation: nonlinear feature drift.
  The degradation mainly appears under the highest unseen speeds and
  loads. When condition parameters, especially speed, change
  drastically, vibration-signal features do not change linearly: at high
  speed, resonance regions may be excited, fundamentally altering the
  spectral structure, and nonlinear inter-component dynamics may be
  amplified. These high-speed-only features, never present in the
  lower-speed training data, cannot be correctly recognized by the
  model.
\item Incompleteness of condition disentanglement: although the model
  strives to decouple operating conditions, when the data-distribution
  drift is too large and extends beyond the convex hull of the training
  data, the disentanglement may be incomplete. The model learns an
  ``interpolation'' invariance within the known range of condition
  variation, and its extrapolation ability declines for drastic
  out-of-range changes.
\item Robustness of anomaly detection: even when the diagnosis accuracy is
  very low (e.g., HUST RF = 65 Hz, 62.92\%), the anomaly-detection
  accuracy remains relatively high (79.64\%). This shows that the
  model's ability to distinguish healthy from faulty
  states is stronger than its ability to discriminate specific fault
  types, and is more robust to operating-condition changes.
\end{steps}

\begin{longtable}[]{@{}
  >{\raggedright\arraybackslash}p{(\columnwidth - 16\tabcolsep) * \real{0.0726}}
  >{\raggedright\arraybackslash}p{(\columnwidth - 16\tabcolsep) * \real{0.0934}}
  >{\raggedright\arraybackslash}p{(\columnwidth - 16\tabcolsep) * \real{0.1037}}
  >{\raggedright\arraybackslash}p{(\columnwidth - 16\tabcolsep) * \real{0.1037}}
  >{\raggedright\arraybackslash}p{(\columnwidth - 16\tabcolsep) * \real{0.1224}}
  >{\raggedright\arraybackslash}p{(\columnwidth - 16\tabcolsep) * \real{0.1349}}
  >{\raggedright\arraybackslash}p{(\columnwidth - 16\tabcolsep) * \real{0.1224}}
  >{\raggedright\arraybackslash}p{(\columnwidth - 16\tabcolsep) * \real{0.1349}}
  >{\raggedright\arraybackslash}p{(\columnwidth - 16\tabcolsep) * \real{0.1120}}@{}}
\caption{Cross-condition zero-shot generalization results.}\label{tab:8}\\
\toprule\noalign{}
\begin{minipage}[b]{\linewidth}\raggedright
\textbf{No.}
\end{minipage} & \begin{minipage}[b]{\linewidth}\raggedright
\textbf{Dataset}
\end{minipage} & \begin{minipage}[b]{\linewidth}\raggedright
\textbf{RF}
\end{minipage} & \begin{minipage}[b]{\linewidth}\raggedright
\textbf{LD}
\end{minipage} & \begin{minipage}[b]{\linewidth}\raggedright
\textbf{Zero-shot acc.}
\end{minipage} & \begin{minipage}[b]{\linewidth}\raggedright
\textbf{Zero-shot anom. acc.}
\end{minipage} & \begin{minipage}[b]{\linewidth}\raggedright
\textbf{Adapt. acc.}
\end{minipage} & \begin{minipage}[b]{\linewidth}\raggedright
\textbf{Adapt. anom. acc.}
\end{minipage} & \begin{minipage}[b]{\linewidth}\raggedright
\textbf{Weighted F1}
\end{minipage} \\
\midrule\noalign{}
\endhead
\bottomrule\noalign{}
\endlastfoot
1 & \multirow{3}{=}{BDS2} & RF2\_1 & LD2\_2 & 0.6493 & 0.8306 & 0.7278 &
0.8771 & 0.7240 \\*
2 & & RF2\_2 & LD2\_1 & 0.9415 & 0.9471 & 0.9492 & 0.9506 & 0.9493 \\*
3 & & RF2\_2 & LD2\_2 & 0.8934 & 0.9017 & 0.8830 & 0.9010 & 0.8830 \\
\midrule
4 & \multirow{3}{=}{BDS3} & RF3\_1 & --- & 0.8662 & 0.9577 & 0.8944 &
0.9577 & 0.8887 \\*
5 & & RF3\_2 & --- & 0.8239 & 0.8697 & 0.9613 & 0.9718 & 0.9732 \\*
6 & & RF3\_3 & --- & 0.6655 & 0.8310 & 0.6620 & 0.8345 & 0.6742 \\
\midrule
7 & \multirow{5}{=}{BDS4} & RF4\_1 & LD4\_1 & 1.0000 & 1.0000 & 1.0000 &
1.0000 & 1.0000 \\*
8 & & RF4\_1 & LD4\_2 & 1.0000 & 1.0000 & 1.0000 & 1.0000 & 1.0000 \\*
9 & & RF4\_1 & LD4\_3 & 1.0000 & 1.0000 & 1.0000 & 1.0000 & 1.0000 \\*
10 & & RF4\_1 & LD4\_5 & 1.0000 & 1.0000 & 1.0000 & 1.0000 & 1.0000 \\*
11 & & RF4\_1 & LD4\_8 & 0.5449 & 0.5897 & 0.3910 & 0.4679 & 0.3504 \\
\midrule
12 & \multirow{3}{=}{BDS5} & --- & LD5\_1 & 1.0000 & 1.0000 & 0.9633 &
0.9633 & 0.9813 \\*
13 & & --- & LD5\_9 & 0.9255 & 0.9255 & 0.9504 & 0.9504 & 0.9745 \\*
14 & & --- & LD5\_38 & 0.9869 & 0.9869 & 0.9869 & 0.9869 & 0.9934 \\
\midrule
15 & \multirow{3}{=}{BDS6} & RF6\_1 & LD6\_3 & 0.7583 & 0.8417 & 0.7500
& 0.8375 & 0.7226 \\*
16 & & RF6\_2 & LD6\_2 & 0.5357 & 0.8393 & 0.6321 & 0.9357 & 0.6011 \\*
17 & & RF6\_3 & LD6\_1 & 0.9667 & 1.0000 & 0.9833 & 1.0000 & 0.9833 \\
\midrule
18 & \multirow{4}{=}{BDS8} & RF8\_1 & --- & 0.8784 & 0.9574 & 0.8055 &
0.8754 & 0.8091 \\*
19 & & RF8\_3 & --- & 0.7720 & 0.9696 & 0.7568 & 0.9514 & 0.7530 \\*
20 & & RF8\_5 & --- & 0.5684 & 0.8419 & 0.6596 & 0.8419 & 0.6122 \\*
21 & & RF8\_7 & --- & 0.6292 & 0.7964 & 0.6261 & 0.7812 & 0.6108 \\
\midrule
22 & \multirow{5}{=}{BDS10} & RF10\_1 & LD10\_1 & 0.5850 & 0.9450 &
0.6200 & 0.9400 & 0.5795 \\*
23 & & RF10\_2 & LD10\_3 & 0.9950 & 1.0000 & 0.9900 & 1.0000 & 0.9900 \\*
24 & & RF10\_3 & LD10\_4 & 1.0000 & 1.0000 & 1.0000 & 1.0000 & 1.0000 \\*
25 & & RF10\_5 & LD10\_2 & 0.9950 & 1.0000 & 0.9950 & 1.0000 & 0.9950 \\*
26 & & RF10\_6 & LD10\_4 & 1.0000 & 1.0000 & 1.0000 & 1.0000 & 1.0000 \\
\end{longtable}

In summary, the foundation model exhibits a useful degree of zero-shot
cross-condition generalization, performing well for interpolated and
mildly varying conditions but degrading for distant extrapolated
conditions. Therefore, introducing new conditions at the
high-speed/heavy-load edge of the range may require modest few-shot
fine-tuning.

\subsection{Task 3: Cross-machine zero-shot generalization}\label{sec:5.5}

\textbf{(1) Task objective.} Evaluate whether a model trained on source
machines (datasets) can generalize well to an unseen target machine
(dataset) with a different structure and data distribution.

\textbf{(2) Task setup.} A leave-one-dataset-out scheme is used: in
turn, one complete dataset is taken as the target test set and the
remaining nine as the source-domain training set. After source-domain
pre-training, the model is evaluated directly on the target test set
without any adjustment.

\begin{figure}[htbp]
\centering
\includegraphics[width=0.85\textwidth]{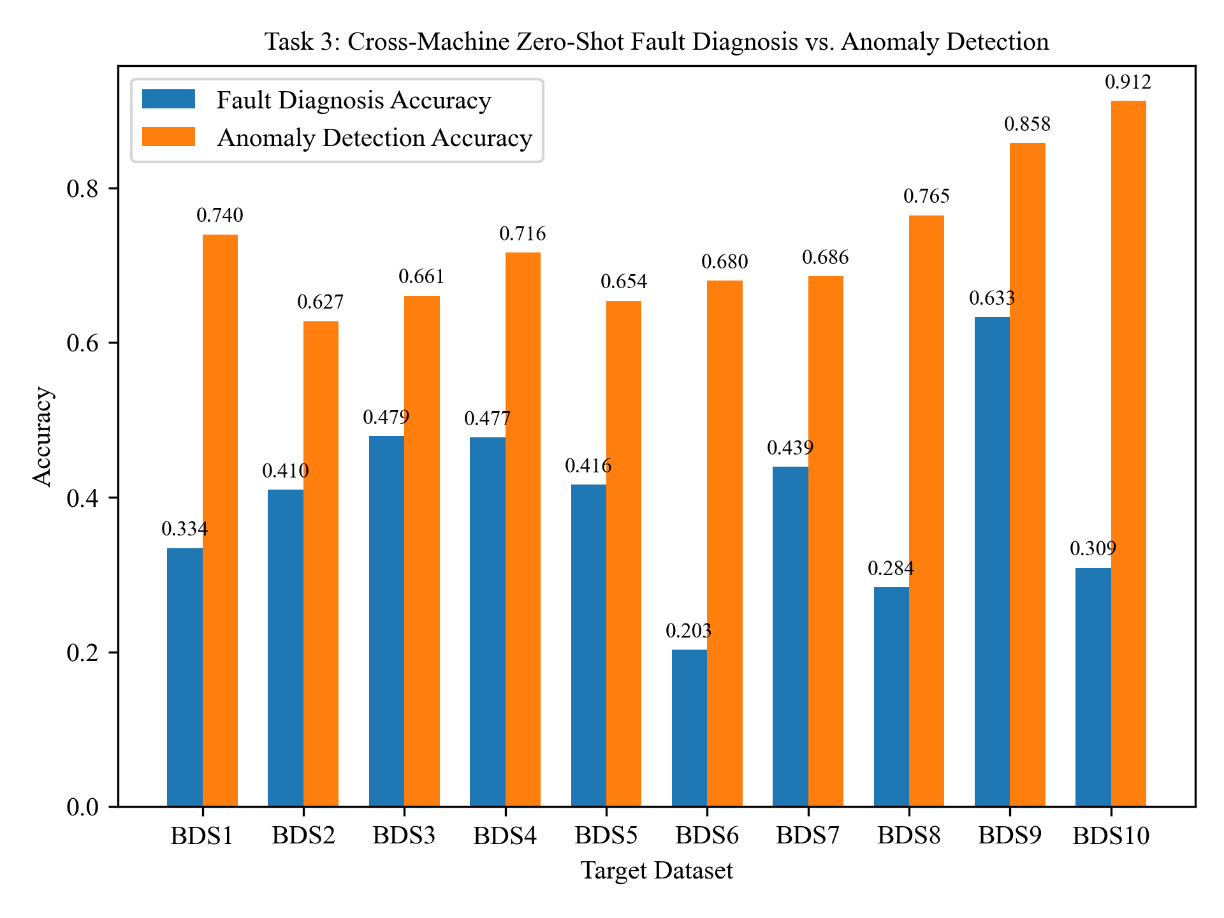}
\caption{Cross-machine zero-shot generalization accuracy.}\label{fig:10}
\end{figure}

\textbf{(3) Result analysis.} The Task-3 results are given in Table~\ref{tab:9}
and Fig.~\ref{fig:10}. The main observations are:

\begin{steps}
\item The model does not yet provide reliable zero-shot cross-machine
  diagnosis: except for BDS9 (SQV, 63.32\%), zero-shot accuracy is below
  50\% in all experiments, showing that directly diagnosing a brand-new
  machine with a model trained on a population of other machines is
  extremely difficult. Compared with condition variation, inter-machine
  differences are a more fundamental domain gap: not only speed and load
  but also bearing model, physical size, sensor type, mounting position,
  transmission path, and background noise all change. Because the
  datasets encountered during pre-training are limited, these systematic
  differences prevent reliable cross-machine inference.
\item Anomaly detection retains a degree of robustness: consistent with Task
  2, although the multi-class diagnosis performance is poor, binary
  anomaly-detection accuracy generally lies between 60\% and 90\%. This
  indicates that the model has learned partially transferable general
  knowledge, namely a universal machine-health representation that
  captures common traits of healthy signals (smooth, impact-free), so
  that when a new machine becomes abnormal and its signal deviates from
  this universal health pattern, the model can still detect it to some
  extent.
\item Analysis of the BDS9 special case: the relatively good zero-shot
  performance on BDS9 (SQV) may reflect similarities in bearing type
  (deep-groove ball bearing), operating condition (50 Hz), and fault
  characteristics between BDS9 and certain training datasets (e.g., the
  50 Hz condition of BDS10 and the deep-groove ball bearings of BDS1 and
  BDS2), so the domain gap is relatively small. Expanding object-type
  coverage and scaling up pre-training therefore offer a promising route
  toward cross-machine zero-shot generalization.
\end{steps}

\begin{longtable}[]{@{}
  >{\raggedright\arraybackslash}p{(\columnwidth - 8\tabcolsep) * \real{0.1709}}
  >{\raggedright\arraybackslash}p{(\columnwidth - 8\tabcolsep) * \real{0.1923}}
  >{\raggedright\arraybackslash}p{(\columnwidth - 8\tabcolsep) * \real{0.2137}}
  >{\raggedright\arraybackslash}p{(\columnwidth - 8\tabcolsep) * \real{0.2308}}
  >{\raggedright\arraybackslash}p{(\columnwidth - 8\tabcolsep) * \real{0.1923}}@{}}
\caption{Cross-machine zero-shot generalization results.}\label{tab:9}\\
\toprule\noalign{}
\begin{minipage}[b]{\linewidth}\raggedright
\textbf{No.}
\end{minipage} & \begin{minipage}[b]{\linewidth}\raggedright
\textbf{Target dataset}
\end{minipage} & \begin{minipage}[b]{\linewidth}\raggedright
\textbf{Zero-shot acc.}
\end{minipage} & \begin{minipage}[b]{\linewidth}\raggedright
\textbf{Zero-shot anom. acc.}
\end{minipage} & \begin{minipage}[b]{\linewidth}\raggedright
\textbf{Weighted F1}
\end{minipage} \\
\midrule\noalign{}
\endhead
\bottomrule\noalign{}
\endlastfoot
3-1 & BDS1 & 0.3341 & 0.7399 & 0.3555 \\*
3-2 & BDS2 & 0.4097 & 0.6275 & 0.3631 \\*
3-3 & BDS3 & 0.4789 & 0.6608 & 0.4987 \\*
3-4 & BDS4 & 0.4774 & 0.7161 & 0.4043 \\*
3-5 & BDS5 & 0.4163 & 0.6539 & 0.3543 \\*
3-6 & BDS6 & 0.2026 & 0.6802 & 0.2695 \\*
3-7 & BDS7 & 0.4392 & 0.6858 & 0.4176 \\*
3-8 & BDS8 & 0.2836 & 0.7647 & 0.3061 \\*
3-9 & BDS9 & 0.6332 & 0.8581 & 0.5890 \\*
3-10 & BDS10 & 0.3088 & 0.9124 & 0.2981 \\
\end{longtable}

\subsection{Task 4: Cross-machine few-shot fine-tuning}\label{sec:5.6}

\textbf{(1) Task objective.} Evaluate whether the pre-trained model can
adapt efficiently to an unseen machine through few-shot fine-tuning with
a small labeled target-machine dataset.

\textbf{(2) Task setup.} Building on Task 3, a very small fine-tuning
set is drawn from the target test dataset by stratified sampling that
jointly considers fault class and operating condition: M samples are
taken for each fault class under at most N condition combinations; the
remaining target samples are used for evaluation. Two few-shot settings
are used: 2-condition/3-sample and 3-condition/5-sample. For example,
2-condition/3-sample means three samples are taken for each fault mode
under each condition, and for datasets with many condition combinations
the number of sampled condition combinations is capped at two.

\begin{figure}[htbp]
\centering
\includegraphics[width=0.91\textwidth]{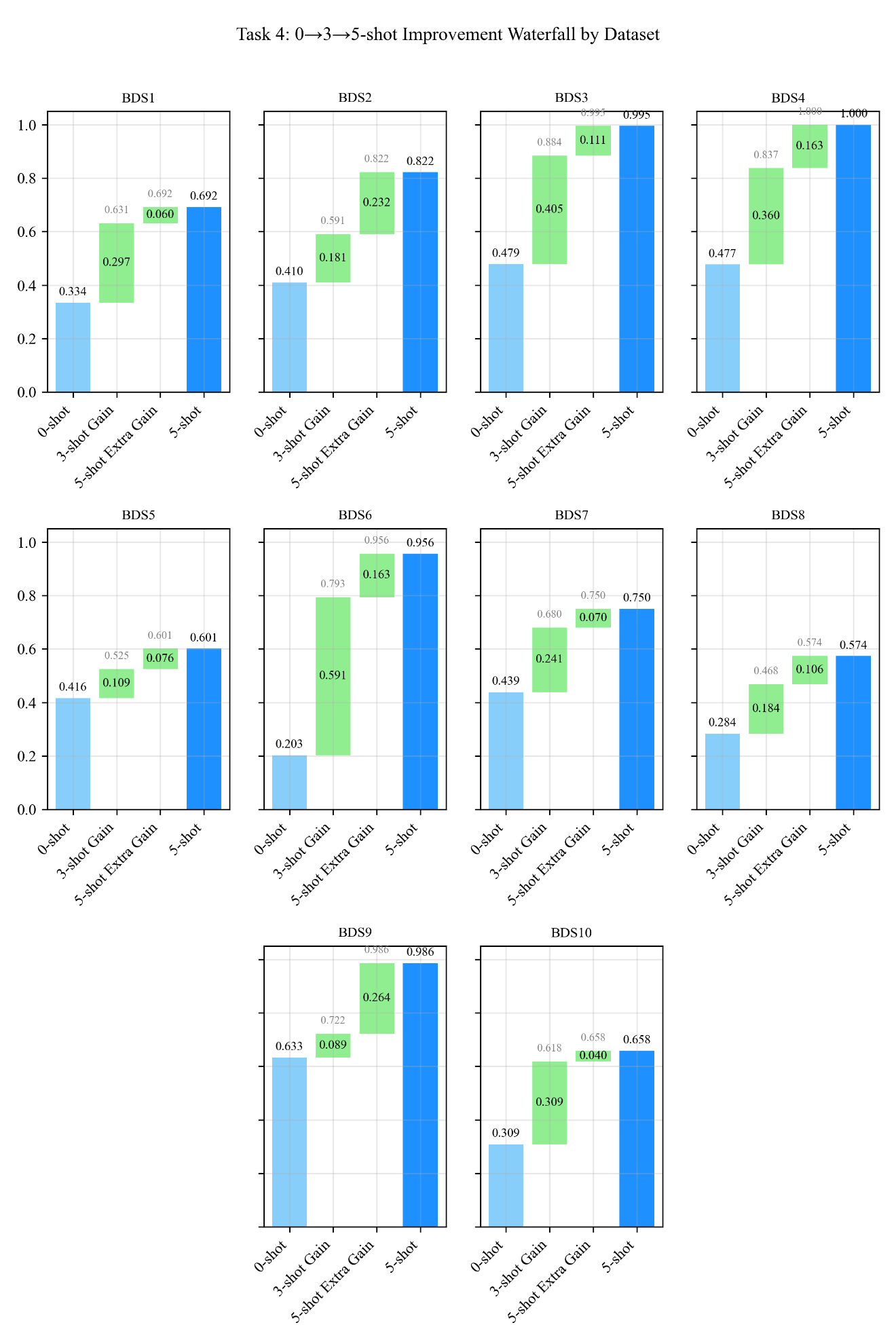}
\caption{Cross-machine zero-shot vs. few-shot accuracy (0\(\rightarrow\)3\(\rightarrow\)5-shot waterfall).}\label{fig:11}
\end{figure}

\textbf{(3) Result analysis.} The Task-4 results are given in Table~\ref{tab:10}
and Fig.~\ref{fig:11}. The main observations are:

\begin{steps}
\item The model has efficient few-shot fine-tuning ability: in nearly all
  experiments, fine-tuning with only 3 to 5 samples per class
  substantially improves diagnosis accuracy. On BDS6 (XJTU) the accuracy
  jumps from 20.26\% to 95.63\%, while on BDS3 and BDS4 it reaches
  99.5\% and 100\%. Such large performance gains under few-shot
  conditions confirm the model's practical potential as
  a foundation model.
\item The model learns transferable fault representations: through
  large-scale pre-training on nine diverse datasets, the model learns
  representations of bearing faults that transfer across datasets,
  meaning its network parameters encode low-level, abstract, and
  transferable feature-analysis capabilities. When applied to a new
  machine, the model need not learn from scratch. It aligns the
  pre-trained representations with the target domain through few-shot
  fine-tuning, using the few new-machine samples as references for
  mapping the existing universal knowledge onto the machine-specific
  fault representations.
\item Few-shot performance is limited in some scenarios: on BDS5 and BDS8,
  the improvement is limited even after 5-shot fine-tuning. This is
  mainly because these datasets contain a large number of condition
  combinations while the experiment caps the number of sampled condition
  combinations at three, which restricts both the number of fine-tuning
  samples and the condition coverage. Therefore, the performance remains
  constrained when the model faces the full set of conditions.
\end{steps}

The results of Tasks 3 and 4 show that direct zero-shot diagnosis on an
unseen machine remains unreliable. However, pre-training on multiple
source datasets followed by few-shot fine-tuning provides effective
cross-machine adaptation. The pre-trained model learns transferable
feature representations, while a small target-machine dataset aligns
those representations with the new domain. This workflow reduces the
data-collection and labeling requirements for deployment on a new
machine.

\begin{longtable}[]{@{}
  >{\raggedright\arraybackslash}p{(\columnwidth - 14\tabcolsep) * \real{0.0833}}
  >{\raggedright\arraybackslash}p{(\columnwidth - 14\tabcolsep) * \real{0.1261}}
  >{\raggedright\arraybackslash}p{(\columnwidth - 14\tabcolsep) * \real{0.1261}}
  >{\raggedright\arraybackslash}p{(\columnwidth - 14\tabcolsep) * \real{0.1389}}
  >{\raggedright\arraybackslash}p{(\columnwidth - 14\tabcolsep) * \real{0.1325}}
  >{\raggedright\arraybackslash}p{(\columnwidth - 14\tabcolsep) * \real{0.1261}}
  >{\raggedright\arraybackslash}p{(\columnwidth - 14\tabcolsep) * \real{0.1389}}
  >{\raggedright\arraybackslash}p{(\columnwidth - 14\tabcolsep) * \real{0.1282}}@{}}
\caption{Cross-machine few-shot fine-tuning results.}\label{tab:10}\\
\toprule\noalign{}
\begin{minipage}[b]{\linewidth}\raggedright
\textbf{No.}
\end{minipage} & \begin{minipage}[b]{\linewidth}\raggedright
\textbf{Target}
\end{minipage} & \begin{minipage}[b]{\linewidth}\raggedright
\textbf{3-shot acc.}
\end{minipage} & \begin{minipage}[b]{\linewidth}\raggedright
\textbf{3-shot anom.}
\end{minipage} & \begin{minipage}[b]{\linewidth}\raggedright
\textbf{3-shot F1}
\end{minipage} & \begin{minipage}[b]{\linewidth}\raggedright
\textbf{5-shot acc.}
\end{minipage} & \begin{minipage}[b]{\linewidth}\raggedright
\textbf{5-shot anom.}
\end{minipage} & \begin{minipage}[b]{\linewidth}\raggedright
\textbf{5-shot F1}
\end{minipage} \\
\midrule\noalign{}
\endhead
\bottomrule\noalign{}
\endlastfoot
4-1 & BDS1 & 0.6315 & 0.9744 & 0.6345 & 0.6917 & 0.9994 & 0.6925 \\*
4-2 & BDS2 & 0.5908 & 0.8250 & 0.5975 & 0.8224 & 0.8830 & 0.8200 \\*
4-3 & BDS3 & 0.8841 & 0.9964 & 0.8809 & 0.9949 & 0.9949 & 0.9950 \\*
4-4 & BDS4 & 0.8373 & 0.9119 & 0.8332 & 1.0000 & 1.0000 & 1.0000 \\*
4-5 & BDS5 & 0.5252 & 0.7815 & 0.5459 & 0.6014 & 0.7682 & 0.6180 \\*
4-6 & BDS6 & 0.7935 & 0.9796 & 0.8055 & 0.9563 & 0.9944 & 0.9567 \\*
4-7 & BDS7 & 0.6801 & 0.8013 & 0.6958 & 0.7500 & 0.8271 & 0.7664 \\*
4-8 & BDS8 & 0.4678 & 0.7657 & 0.4707 & 0.5739 & 0.8346 & 0.5725 \\*
4-9 & BDS9 & 0.7221 & 0.7502 & 0.7145 & 0.9863 & 0.9911 & 0.9867 \\*
4-10 & BDS10 & 0.6183 & 0.9509 & 0.6055 & 0.6578 & 0.8987 & 0.6561 \\
\end{longtable}

\subsection{Comparative analysis against domain-generalization baselines and a bearing foundation model}\label{sec:5.7}

To compare the generalization and few-shot performance of the proposed
model, two recent publicly available studies are selected.

\textbf{(1) Cross-condition generalization comparison.} For
cross-condition diagnosis, the main comparison is against a
comprehensive cross-domain benchmark study~\cite{ref13}, whose purpose is to
fairly and independently reproduce and evaluate a series of existing
methods under unified public datasets and evaluation standards. Both
that study and the present case study adopt the PU dataset with highly
consistent data selection and cross-condition settings, making the
comparison meaningful. The benchmark evaluates eight popular methods
covering classical and state-of-the-art domain-generalization models:
empirical risk minimization (ERM)~\cite{ref57}, domain-adversarial neural
network (DANN)~\cite{ref12}, deep correlation alignment (DCORAL)~\cite{ref58},
deep domain confusion (DDC)~\cite{ref59}, CNN with center loss (CNN-C)~\cite{ref60}, the intrinsic-and-extrinsic domain-generalization network
(IEDGNet)~\cite{ref61}, the conditional contrastive domain-generalization
network (CCDG)~\cite{ref62}, and the domain-generalization network combining
invariance and specificity (DGNIS)~\cite{ref20}. It should be noted that
this paper divides PU into three condition combinations, whereas the
benchmark further subdivides one of them into two (four conditions in
total); the present cross-condition span is therefore larger, and for a
fair comparison the two subdivided conditions of the benchmark are
averaged before comparison.

\begin{longtable}[]{@{}
  >{\raggedright\arraybackslash}p{(\columnwidth - 8\tabcolsep) * \real{0.1761}}
  >{\raggedright\arraybackslash}p{(\columnwidth - 8\tabcolsep) * \real{0.2201}}
  >{\raggedright\arraybackslash}p{(\columnwidth - 8\tabcolsep) * \real{0.2201}}
  >{\raggedright\arraybackslash}p{(\columnwidth - 8\tabcolsep) * \real{0.2201}}
  >{\raggedright\arraybackslash}p{(\columnwidth - 8\tabcolsep) * \real{0.1635}}@{}}
\caption{Cross-condition generalization comparison on the PU dataset (accuracy, \%).}\label{tab:11}\\
\toprule\noalign{}
\begin{minipage}[b]{\linewidth}\raggedright
\textbf{Method}
\end{minipage} & \begin{minipage}[b]{\linewidth}\raggedright
\textbf{RF2\_1 \& LD2\_2}
\end{minipage} & \begin{minipage}[b]{\linewidth}\raggedright
\textbf{RF2\_2 \& LD2\_2 (mean)}
\end{minipage} & \begin{minipage}[b]{\linewidth}\raggedright
\textbf{RF2\_2 \& LD2\_1}
\end{minipage} & \begin{minipage}[b]{\linewidth}\raggedright
\textbf{Average}
\end{minipage} \\
\midrule\noalign{}
\endhead
\bottomrule\noalign{}
\endlastfoot
ERM & 32.47 & 76.94 & 98.65 & 71.24 \\*
DDC & 31.77 & 75.11 & 97.75 & 69.93 \\*
DANN & 38.76 & 75.04 & 98.19 & 71.75 \\*
DCORAL & 39.45 & 75.24 & 99.04 & 72.24 \\*
CNN-C & 34.72 & 73.86 & 98.35 & 70.20 \\*
IEDGNet & 31.89 & 76.08 & 95.40 & 69.86 \\*
CCDG & 28.17 & 76.67 & 97.72 & 69.81 \\*
DGNIS & 25.65 & 66.03 & 70.22 & 56.98 \\*
TFC-MMFD (ours) & 64.93 & 89.34 & 94.15 & 82.81 \\
\end{longtable}

\begin{figure}[htbp]
\centering
\includegraphics[width=0.66\textwidth]{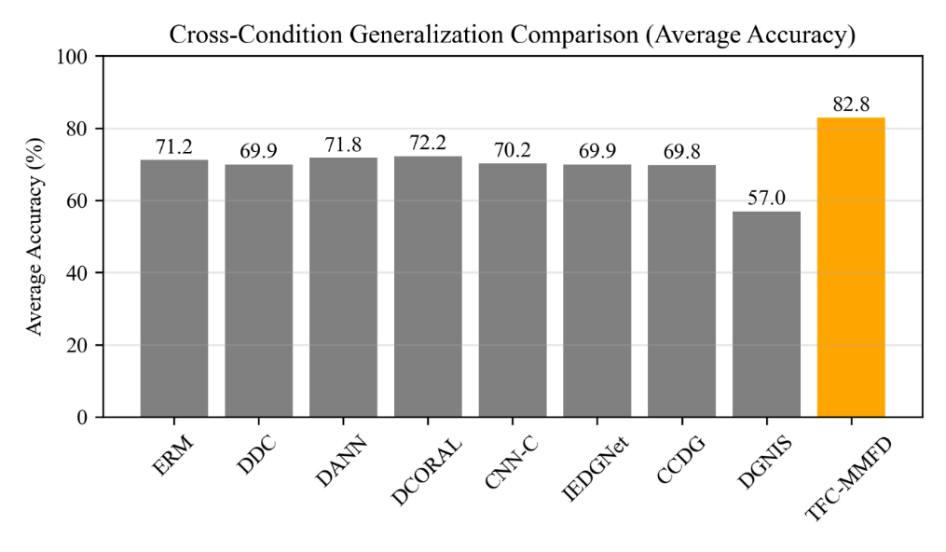}
\caption{Average cross-condition generalization accuracy on the PU dataset.}\label{fig:12}
\end{figure}

On PU cross-condition diagnosis (Table~\ref{tab:11} and Fig.~\ref{fig:12}), traditional
domain-generalization and transfer methods cluster within 69-72\% (best
baseline DCORAL at 72.24\%), while DGNIS (56.98\%) highlights the limits
of classical statistical matching under highly differentiated
conditions. TFC-MMFD reaches 82.81\% average accuracy, which is 10.57
percentage points above DCORAL (about 14.6\% relative improvement), and
does so under a larger cross-domain span (three coarser conditions here
versus four finer ones in the benchmark). The advantage likely stems
from mechanism differences: whereas traditional methods rely on
adversarial alignment or statistical-distribution matching that struggle
with wide fault bands and complex temporal structure, TFC-MMFD aligns
key band energies and time-frequency patterns across conditions via
contrastive and adversarial training, and embeds raw signals,
time-frequency images, and semantic labels into a unified space,
suppressing condition-specific noise and learning condition-independent
yet fault-relevant high-level semantics.

\begin{figure}[htbp]
\centering
\includegraphics[width=1.00\textwidth]{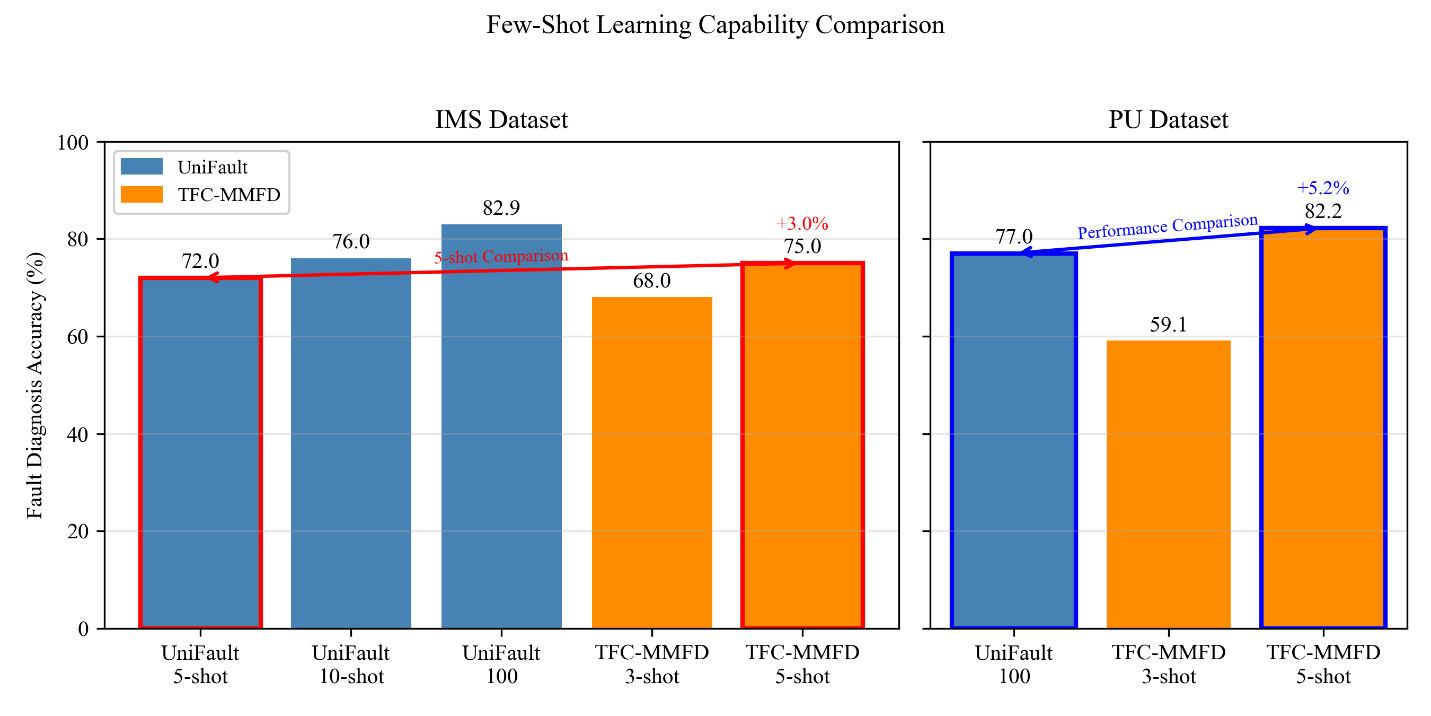}
\caption{Few-shot diagnosis accuracy: TFC-MMFD vs. UniFault.}\label{fig:13}
\end{figure}

\textbf{(2) Few-shot learning comparison.} For few-shot learning, the
main comparison is against UniFault~\cite{ref63}, a recent bearing-data
fault-diagnosis foundation model jointly published by researchers from
A*STAR and Nanyang Technological University with Prof. Jay Lee of the
University of Maryland. UniFault is a Transformer-based foundation model
pre-trained on a large-scale database of about 9 billion data points
spanning ten heterogeneous bearing datasets (IMS, UO, CWRU, PU, Torino,
XJTU-SY, MFPT, FEMTO, KAIST, HIT-SM), and its case validation evaluates
few-shot fine-tuning performance on three target datasets (IMS, UO, PU)
with the few-shot count set to 100 samples. Here the comparison uses the
overlapping datasets IMS and PU. In TFC-MMFD, ``3-shot'' and ``5-shot''
denote samples per fault per condition (with the number of sampled
condition combinations capped at 2 and 3, respectively), so the totals
are far fewer than UniFault's 100; furthermore,
UniFault's few-shot samples are themselves drawn from
its pre-training data, i.e., they are not entirely new, whereas TFC-MMFD
uses no target-dataset samples during pre-training and only genuinely
new data during few-shot fine-tuning. As Table~\ref{tab:12} and Fig.~\ref{fig:13} show, on
PU, TFC-MMFD reaches 82.24\% accuracy with 45 fine-tuning samples,
exceeding UniFault's 77.02\% with 100 samples; on IMS,
TFC-MMFD also shows a slight edge in the few-shot setting, attaining
75.00\% under 5-shot, slightly above UniFault's
estimated \textasciitilde72\% under comparable conditions. These
comparisons indicate that TFC-MMFD uses the available target data
efficiently to capture the fault characteristics of a new machine from a
small target dataset.

\begin{longtable}[]{@{}
  >{\raggedright\arraybackslash}p{(\columnwidth - 8\tabcolsep) * \real{0.1496}}
  >{\raggedright\arraybackslash}p{(\columnwidth - 8\tabcolsep) * \real{0.2030}}
  >{\raggedright\arraybackslash}p{(\columnwidth - 8\tabcolsep) * \real{0.2350}}
  >{\raggedright\arraybackslash}p{(\columnwidth - 8\tabcolsep) * \real{0.2062}}
  >{\raggedright\arraybackslash}p{(\columnwidth - 8\tabcolsep) * \real{0.2062}}@{}}
\caption{Few-shot learning comparison between TFC-MMFD and UniFault.}\label{tab:12}\\
\toprule\noalign{}
\begin{minipage}[b]{\linewidth}\raggedright
\textbf{Dataset}
\end{minipage} & \begin{minipage}[b]{\linewidth}\raggedright
\textbf{Model}
\end{minipage} & \begin{minipage}[b]{\linewidth}\raggedright
\textbf{Fine-tune samples}
\end{minipage} & \begin{minipage}[b]{\linewidth}\raggedright
\textbf{Accuracy}
\end{minipage} & \begin{minipage}[b]{\linewidth}\raggedright
\textbf{F1}
\end{minipage} \\
\midrule\noalign{}
\endhead
\bottomrule\noalign{}
\endlastfoot
\multirow{5}{=}{IMS} & UniFault & 5-shot & \textasciitilde72 (est.) &
--- \\*
& UniFault & 10-shot & \textasciitilde76 (est.) & --- \\*
& UniFault & 100 samples & 82.94 & 82.09 \\*
& TFC-MMFD & 3-shot (12 total) & 68.01 & 69.58 \\*
& TFC-MMFD & 5-shot (20 total) & 75.00 & 76.64 \\
\midrule
\multirow{3}{=}{PU} & UniFault & 100 samples & 77.02 & 75.82 \\*
& TFC-MMFD & 3-shot (27 total) & 59.08 & 59.75 \\*
& TFC-MMFD & 5-shot (45 total) & 82.24 & 82.00 \\
\end{longtable}

\subsection{Summary of the case study}\label{sec:5.8}

Through the four progressive tasks, the base performance,
generalization, and rapid few-shot adaptation potential of TFC-MMFD are
evaluated:

\begin{enumerate}
\item \textbf{Universal diagnosis under heterogeneous conditions:} under
  ideal data conditions, the model achieves reliable performance
  (99.40\% accuracy), confirming the effectiveness of the multimodal
  fusion architecture and LLaMA backbone for unified modeling of
  heterogeneous objects, conditions, and data.
\item \textbf{Cross-condition generalization:} the model generalizes
  zero-shot to unseen conditions within a range, particularly
  interpolated conditions; extrapolated conditions far from the training
  distribution, especially high-speed, heavy-load conditions, still
  require improvement.
\item \textbf{Cross-machine adaptation as a foundation model:} although
  zero-shot cross-machine diagnosis remains unreliable, the pre-trained
  model has practical value. Learning general knowledge on source
  domains and applying parameter-efficient few-shot fine-tuning on
  target machines enables rapid, low-cost adaptation with high accuracy,
  reducing the data and development requirements for deployment on new
  machines.
\end{enumerate}

Overall, the proposed TFC multimodal fusion foundation model and its
``large-scale pre-training + few-shot fine-tuning'' paradigm demonstrate
the feasibility and generalization of unified modeling and universal
diagnosis, providing a possible solution to the long-standing challenges
of condition shift and machine difference in industrial fault diagnosis.

\section{Conclusions}\label{sec:6}

To overcome the strong task-specificity and weak generalization of
conventional diagnosis models under multi-element heterogeneous
scenarios, this paper proposed the Time-Frequency-Condition Multimodal
Fault Diagnosis foundation model (TFC-MMFD). Within an end-to-end
framework, heterogeneous time-, frequency-, and condition-domain inputs
are first tokenized into a unified language-like space. A LLaMA-centered
architecture comprising multimodal space mapping, bidirectional
cross-attention, and Q-Former query generation then progressively
refines these signals into discriminative fault representations. By
combining regression-adversarial condition disentanglement, masked
self-supervision, and fault-mode-anchor matching, TFC-MMFD attains
cross-condition generalization and rapid cross-machine adaptation.
Extensive experiments on ten public bearing datasets under four
diagnostic tasks lead to the following conclusions:

\textbf{(1) The tokenized unified language-like representation and the
hierarchical fusion architecture make unified modeling of multi-source
heterogeneous data feasible and effective.} On the mixed benchmark
spanning ten datasets with diverse bearing types, speeds, loads, and
sampling rates, the model achieved 99.40\% overall accuracy, a 99.16\%
macro-average F1, and 99.77\% anomaly-detection accuracy; the
near-identical macro- and weighted-average F1 values indicate unbiased
recognition under severe class imbalance, and t-SNE visualization
illustrates a layer-by-layer progression from raw signals to
discriminative feature representations.

\textbf{(2) The regression-adversarial and condition-conditioned
training mechanisms deliver reliable zero-shot generalization within the
training condition envelope.} Under leave-one-condition-out evaluation,
zero-shot accuracy exceeded 80\% in most trials and reached 100\% for
interpolated conditions, and the average cross-condition accuracy on the
PU dataset surpassed the best of eight representative
domain-generalization baselines by 10.57 percentage points. Performance
degraded for extrapolated boundary conditions (e.g., 87.84\% at the
lowest versus 62.92\% at the highest speed on the HUST dataset),
delineating an interpolation boundary of the learned condition
invariance that is attributable to nonlinear feature drift outside the
training support.

\textbf{(3) The fault-mode-anchor matching mechanism provides
interpretable diagnosis and transferable health knowledge across
machines.} Although zero-shot multi-class accuracy on entirely new
machines stayed mostly below 50\%, which reflects the fundamental domain
gap among machines, binary anomaly detection remained at 60-90\%,
showing that the anchor-based approach captures a general representation
of machine health that transfers across objects. Moreover, because new
fault modes enter the output space by simply adding their text anchors,
the diagnostic scope can be extended without retraining a fixed
classification head.

\textbf{(4) The pre-training-fine-tuning paradigm is a practical and
efficient route to rapid new-machine deployment.} In the three- and
five-shot settings, accuracy on new machines increased substantially
(e.g., from 20.26\% to 95.63\% on the XJTU dataset), and the model
outperformed the UniFault bearing foundation model by 5.2 percentage
points on the PU dataset while using less than half of its fine-tuning
data, none of which appeared during pre-training. This indicates that
the model has learned transferable, low-level representations of bearing
faults.

Several limitations should be acknowledged. First, generalization
degrades for operating conditions extrapolated beyond the training
range, because the nonlinear spectral changes induced by extreme speeds
and loads are absent from the training distribution. Second, the
pre-training data are insufficient for reliable zero-shot cross-machine
diagnosis. Third, fault modes with scarce samples are still occasionally
confused with adjacent modes in the feature space. Finally, although the
TFC framework is intended for broader equipment and sensor types, the
current evaluation is confined to single-channel vibration signals from
bearing datasets.

Future work will extend the model to other rotating components and to
reciprocating machinery, hydraulic systems, and power-electronic
equipment. Additional sensor signals, including temperature, pressure,
and acoustic emission, will be incorporated to support diagnosis and
prognosis. Physics-informed priors and causal-inference methods will
also be used to improve model performance under extreme extrapolated
conditions and out-of-distribution generalization. Finally, larger and
more carefully annotated industrial datasets will be constructed to
support future fault-diagnosis foundation models.

\section*{Declaration of competing interest}

The authors declare that they have no known competing financial
interests or personal relationships that could have appeared to
influence the work reported in this paper.

\section*{Data availability}

All ten bearing datasets analyzed in this study are publicly available
from the sources cited in Section~\ref{sec:5.1}. The processed data and code are available from the corresponding author upon reasonable request.

\section*{Acknowledgments}

This work was supported by the National Key R\&D Program of China under Grant STI 2030-Major Projects No.2021ZD0201300 and by the Fundamental Research Funds for the Central Universities.

\end{document}